\documentclass{article}

\usepackage{microtype}
\usepackage{graphicx}
\usepackage{subcaption} 
\usepackage{booktabs} 
\usepackage{listings}

\usepackage[colorlinks=true, citecolor=blue]{hyperref}

\usepackage{amsmath,amsfonts,bm, amsthm}

\def\eqref#1{equation~\ref{#1}}

\def\1{\bm{1}}

\def\rvx{{\mathbf{x}}}

\def\rvz{{\mathbf{z}}}

\def\vmu{{\bm{\mu}}}

\def\evsigma{{\sigma}}

\def\mSigma{{\bm{\Sigma}}}

\DeclareMathAlphabet{\mathsfit}{\encodingdefault}{\sfdefault}{m}{sl}
\SetMathAlphabet{\mathsfit}{bold}{\encodingdefault}{\sfdefault}{bx}{n}

\newcommand{\E}{\mathbb{E}}
\newcommand{\Pdist}{\mathbb{P}}
\newcommand{\Qdist}{\mathbb{Q}}

\newcommand{\R}{\mathbb{R}}

\newtheorem*{theorem*}{Theorem}
\newtheorem{theorem}{Theorem}
\newtheorem{definition}{Definition}
\newtheorem*{proposition*}{Proposition}
\newtheorem{proposition}{Proposition}

\usepackage[accepted]{icml2020}

\icmltitlerunning{Revisiting Factorizing Aggregated Posterior}

\begin{document}

\twocolumn[
\icmltitle{Revisiting Factorizing Aggregated Posterior \\ in Learning Disentangled Representations}



\icmlsetsymbol{equal}{*}

\begin{icmlauthorlist}
\icmlauthor{Ze Cheng}{1,2}
\icmlauthor{Juncheng Li}{2,3}
\icmlauthor{Chenxu Wang}{1}
\icmlauthor{Jixuan Gu}{to,go}
\icmlauthor{Hao Xu}{to,5}
\icmlauthor{Xinjian Li}{3}
\icmlauthor{Florian Metze}{3}
\end{icmlauthorlist}

\icmlaffiliation{1}{Bosch China Investment Ltd}
\icmlaffiliation{2}{Bosch Center for AI}
\icmlaffiliation{3}{School of Computer Science, Carnegie Mellon University}
\icmlaffiliation{go}{School of Mathematical Sciences, Shanghai Jiaotong University}
\icmlaffiliation{5}{Department of Applied Mathematics, University of Colorado at Boulder}
\icmlaffiliation{to}{Work as intern in Bosch China}

\icmlcorrespondingauthor{Ze Cheng}{ze.cheng@cn.bosch.com}

\icmlkeywords{Total Correlation, Representative Learning, Disentanglement}
\vskip 0.3in
]



\printAffiliationsAndNotice{}  

\begin{abstract}

In the problem of learning disentangled representations, one of the promising methods is to factorize aggregated posterior by penalizing the total correlation of \textit{sampled} latent variables. However, this well-motivated strategy has a blind spot: there is a disparity between the \textit{sampled} latent representation and its corresponding \textit{mean} representation.
In this paper, we provide a theoretical explanation that low total correlation of \textit{sampled} representation cannot guarantee low total correlation of the \textit{mean} representation.
Indeed, we prove that for the multivariate normal distributions, the \textit{mean} representation with arbitrarily high total correlation can have a corresponding sampled representation with bounded total correlation.
We also propose a method to eliminate the above-mentioned disparity. 
Experiments show that our model can learn a \textit{mean} representation with much lower total correlation, hence a factorized mean representation. 
Moreover, we offer a detailed explanation of the limitations of factorizing aggregated posterior: factor disintegration. Our work indicates a potential direction for future research of disentangled learning. 
\end{abstract}

\textit{Disentangled representation} is believed to be the key to learn a better representation~\citep{bengio2013representation, lecun2015deep, peters2017elements}. 
There are \textit{2 major ingredients of disentanglement:} 1. Models should learn separate \textit{factors of variations} \citep{bengio2013representation}; 2. Factors should be compact \citep{bengio2013representation}, informative and independent from task at hand~\citep{goodfellow2009measuring}.
The motivation of disentanglement includes usefulness for downstream tasks \citep{bengio2013representation}, being invariant to nuisance factors \citep{kumar2017variational}, improving robustness to adversarial attack \citep{alemi2016deep}, etc. (See also the introduction of disentangled representation in \citet{locatello2018challenging, chenisolating, kim2018disentangling} and reference therein.)


Recent works~\citep{higgins2017beta,kim2018disentangling,chenisolating,kumar2017variational, ridgeway2018learning} 
have introduced various regularizers to the objective function of the \textit{Variational Autoencoder} (VAE)~\citep{kingma2013auto, bengio2007scaling}, \textit{Evidence Lower Bound} (ELBO). They aim at factorizing \textit{aggregated posterior}, $q(z)=\int q(z|x)p(x)dx$, which hopefully can encourage disentanglement.
Among these works, \citet{kim2018disentangling,chenisolating} independently proposed a promising regularizer, the total correlation (TC) of sampled representation. TC is defined to be the KL-divergence between the joint distribution $\rvz\sim q(z)$ and the product of marginal distributions $\prod_j q(z_j)$. The TC of a sampled representation, $TC_{sample}$, should describe its level of independence. In this case, a low value suggests a more factorized joint distribution. 

However, \citet{locatello2018challenging} point out, though these works seem to be effective at factorizing aggregated posterior, there exists a blind spot: a disparity between $TC_{sample}$ and the TC of the corresponding mean representation, $TC_{mean}$. Specifically, A low $TC_{sample}$ does not necessarily give rise to a low $TC_{mean}$.
Conventionally, the mean representation is used as the encoded latent variables, an unnoticed high $TC_{mean}$ is usually the culprit behind the undesirable entanglement.
They found that as the strength of regularization on $TC_{sample}$ increases, $TC_{sample}$ decreases as expected, 
but $TC_{mean}$ increases. Moreover, the scores under disentanglement metrics are uncorrelated to the regularization strength.
Their finding has 2 implications: 
\begin{enumerate}
    \item Low $TC_{sample}$ does not imply low $TC_{mean}$, which is yet not understood;
    \item Either low $TC_{sample}$ or low $TC_{mean}$ does not guarantee disentanglement. Disentanglement does not seem to correlate with $TC_{sample}$ or $TC_{mean}$ no matter how much these 2 quantities change.
\end{enumerate}


This created several important yet not answered questions: Why does $TC_{sample}$ have no control over $TC_{mean}$? Is the strategy of regularizing TC unhelpful to disentanglement?
In this paper, we answer the first question completely by theoretically analyzing the relation between $TC_{sample}$ and $TC_{mean}$. Then after investigating factorized representations, we believe that regularizing TC still might be a key to learning disentangled representation and hopefully our study can shed some light into this problem.

Our main contributions are listed as the followings:
\begin{itemize}
    \item 
    We prove that for all mean representations in multivariate normal distribution, there exists a large class of sample distributions with bounded TC
    ~(See Theorem \ref{tcbound}). 
    This implies that a low TC of sample distribution cannot guarantee a low TC of mean representation.  (Section.~\ref{background})
    
    \item 
    We show how to control both $TC_{sample}$ and $TC_{mean}$ and obtain factorized mean representation. Our method is to introduce a simple yet effective regularizer, a penalty term on the \textit{variance} of each latent variable, which forces a sampled representation to behave similarly to the corresponding mean representation. 
    (Section.~\ref{rtcVAE})
    
    
    \item We compare different methods of TC estimation and point out that the method of minibatch estimators (MSS/MWS) suffers from the curse of dimensionality, i.e., the estimation accuracy decays significantly with the increase of the dimension of the latent space. In addition, they may cause unintended shutdown of latent dimensions. 
    (Section.~\ref{sec:TCestimate})
    
    
    
    \item We investigate the limitation of factorized mean representation and suggest a tradeoff be considered for the future work of learning disentangled representation. 
    (Section.~\ref{Experiments})
\end{itemize}

\section{Related Works}
\label{relatedWorks}
VAE~\citep{kingma2013auto, bengio2007scaling} takes the variational approach to approximate the posterior $p(z|x)$ with $q(z|x)$ by minimizing their KL-divergence, $\mathrm{KL}(q(z|x)\|p(z|x))$, which is equivalent to maximizing ELBO.
As a result, the high-dimensional real world observations $\rvx$ is encoded into lower-dimension latent variable $\rvz$ that is expected to be semantically meaningful. 

In order to learn disentangled representation, \citet{higgins2017beta} proposed a modification of the VAE framework and introduced an adjustable hyperparameter $\beta$ that balances latent channel capacity and independence constraints with reconstruction accuracy. 

\citet{chenisolating} proposed $\beta$-TCVAE which adopts the idea of decomposing the average ELBO \citep{hoffman2016elbo} and penalizes the TC of latent variables aiming on regularizing a more precise source of disentanglemnet. Around the same time, \citet{kim2018disentangling} proposed a similar regularizer penalizing $TC_{sample}$ called FactorVAE. The major difference between FactorVAE and $\beta$-TCVAE lies in their different strategies of estimating $TC_{sample}$. \citet{chenisolating} used formulated estimators while \citet{kim2018disentangling} utilized the density-ratio trick which requires an auxiliary discriminator network. 
We will discuss these two strategies more in details in Section.~\ref{sec:TCestimate}.~\citet{kumar2017variational} introduced DIP-VAE-I\&II, which penalize on the covariance matrix of mean and sampled latent variables respectively in order to encourage disentanglement. This strategy could learn an uncorrelated but not independent distribution.

\citet{locatello2018challenging} challenged most recent work on disentanglement and argued that unsupervised learning of disentangled representations without inductive biases is basically impossible. This makes strong suggestion that researchers should pay attention to representative learning with inductive biases on both learning approaches and data sets. We refer readers to works in this direction, e.g. \citet{thomas2018disentangling,bouchacourt2018multi,rolinek2019variational} and works referred therein. However, \citet{locatello2018challenging} does not provide an explanation to one of  the observations they made, i.e., why most regularizers are effective at factorizing aggregated posterior but the corresponding mean representations may be entangled? We answer this question in the next section.

\section{The Disparity between Total Correlation of Mean and Sampled Distribution}
\label{background}
In information theory, total correlation (TC) is one of the generalizations of mutual information (see definition \ref{totalcorrelation}), which measures the difference between the joint distribution of multiple random variables and the product of their marginal distributions. A high value of TC indicates the joint distribution is far from an independent distribution, and hence it suggests high entanglement among these random variables.
\begin{definition}\label{totalcorrelation}
The total correlation of random variable $\rvx$ is defined as
\[\mathrm{TC}(\rvx) := \mathrm{KL}\left( p(x) || \displaystyle \prod_j p(x_j)\right) 
= \E_{p(x)}\left[ \log \frac{p(x)}{\prod_j p(x_j)} \right].\]
\end{definition}
Motivated by this concept, people seek the solution of disentanglement in the form of low TC of the latent variables \citep{kim2018disentangling,chenisolating}. 
However, \citet{locatello2018challenging} pointed out that even though $TC_{sample}$ is low, $TC_{mean}$ can be high. This is problematic because the mean representation is usually taken as the representation of input and such representation is entangled. 

Hence, a clear understanding of the relation between $TC_{sample}$ and $TC_{mean}$ is needed.
To this end, we present a theorem that provides an explicit bound for $TC_{sample}$ under some mild assumptions. This bound does not rely on the distribution of mean representation, which turns out to be the root of the disparity between the two TCs. 
One of the assumptions we made is that the distribution of mean representation is multivariate normal (MVN), but actually the theorem can be easily generalized to distributions with compact support or fast decay. This makes the theorem relatively general and effective for many practical cases.

Here are some notations: $\vmu$ and $\rvz$ are random variables, and $\mu$ and $z$ are corresponding samples (fixed values); $\mSigma$ and $\mSigma'$ are matrices; $C$ stands for some constant.

\begin{figure}

\raisebox{-\height}{\includegraphics[width=0.8\columnwidth]{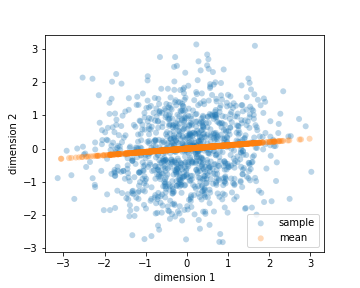}}
\centering
\caption{As an example of disparity between mean and sampled representation, consider the following $\vmu$ and $\rvz$. 
Let $(\vmu_1,\vmu_2)\sim \mathcal{N}(0, \mSigma)$, where $\mSigma = \left(\begin{array}{cc}
         1 & 0.1 \\
         0.1 & 0.01
\end{array} \right)$, then $TC_{mean} = TC(\vmu_1,\vmu_2)=\infty$. Note that $\vmu_1$ has almost shut down. Also, let $\rvz_1|\mu_1\sim\mathcal{N}(\mu_1, 0.01)$ and $\rvz_2|\mu_2\sim\mathcal{N}(\mu_2, 1)$, then $TC_{sample}=TC(\rvz_1,\rvz_2)$ is very low. Such problem (high $TC_{mean}$ and low $TC_{sample}$) exists in $\beta$-TCVAE and FactorVAE.\\}
\label{fig:disparity}
\end{figure}

\begin{theorem}\label{tcbound}
Let $\vmu \sim \mathcal{N}(0, \mSigma)$. 
For a fixed $\mu$, let $\rvz|\mu \sim \mathcal{N}(\mu, \mSigma'(\mu))$, where $\mSigma'(\mu)$ is diagonal and satisfies that,
\begin{align}
    c_1\leq \sigma'_j(\mu) \leq c_2,
\end{align}
where $c_1,c_2>0$ and $\sigma'_j(\mu)$ is the diagonal element of $\mSigma'(\mu)$.
Then $TC(\rvz)$ is independent of $TC(\vmu)$ and 
\begin{align}
    \mathrm{TC}(\rvz) \leq C\frac{c_3^D}{c_1^D} \log \left(\frac{c_2}{c_1} \right) 
            + C\frac{c_3^{D+2}}{c_1^{D+2}},
\end{align}
where $c_3=\max(c_2, \sqrt{D})$ and $C$ is some constant that replies only on dimension $D$.
\end{theorem}
The details of the proof are presented in Appendix \ref{app:tcbound_proof}. Intuitively, in the case of multivariate normal distribution, if there exist two dimensions of $\vmu$ with high correlation, then the TC of this distribution is high (less independent). And the probability density is narrowly distributed in the subspace of these two dimensions. Now, if the standard deviations of $\rvz|\mu$ corresponding to these two dimensions are suitably large (bounded away from zero), then the distribution of $\rvz$ will spread wider in the subspace which is closer to a distribution with low TC (more independent). Figure~\ref{fig:disparity} gives an example, for a distribution of $\vmu$ with high TC, how to construct distribution of $\rvz$ with low TC.

One fact can be deduced from Theorem~\ref{tcbound} is: with a fixed upperbound of $TC_{sample}=TC(\rvz)$ (by fixing parameters $c_1,c_2$), one can make $TC_{mean}=TC(\vmu)$ arbitrarily large. To see this, we use Proposition \ref{tc_normal} in Section.~\ref{sec:TCestimate}, which states $TC(\vmu)$ depends only on the determinant of the correlation matrix of $\vmu$, i.e., $|\mSigma|$, so we only need to tune the off-diagonal elements of $\mSigma$ (while keeping $c_1 ,c_2$ unchanged) to make $|\mSigma|$ go to zero and hence $TC(\vmu)$ go to infinity.

Interestingly, we note that in Theorem \ref{tcbound} when $c_1=c_2$ approaches zero, the upperbound of $TC(\rvz)$ goes to infinity. It reflects the following fact: when the distribution of $\rvz$ is closed to the distribution of $\vmu$, $TC(\rvz)$ is close to $TC(\vmu)$, which can be large. 

Thus, Theorem \ref{tcbound} provides an explanation to the disparity observed by \citet{locatello2018challenging} that $\mathrm{TC}(\rvz)$ is low but $\mathrm{TC}(\vmu)$ is high. 
Indeed, for every distribution of $\vmu$ with large $TC(\vmu)$ there exist a family of distributions of $\rvz|\mu$ with bounded $TC(\rvz)$. If the objective function only penalizes $TC(\rvz)$, the optimization process could easily find a distribution of $(\vmu, \rvz)$ with low $TC(\rvz)$ but high $TC(\vmu)$.
However, this disparity can be eliminated. 
In Section.~\ref{rtcVAE}, we propose a simple regularizer to serve this goal.


\section{An Additional Regularizer}
\label{rtcVAE}
To simplify notation, let $p(n) = p(x_n)$, $q(z|n)=q(z|x_n)$ and $p(n|z)=p(x_n|z)$. Recall the average \textit{evidence lower bound} (ELBO):
\begin{align}\label{ELBO}
    \mathrm{ELBO} := \E_{p(n)}\left[\E_{q(z|n)}[\log p(n|z)] - \mathrm{KL}(q(z|n)\|p(z))\right],
\end{align}
where the first term can be interpreted as \textit{reconstruction error}.
Inspired by ELBO decomposition \citep{hoffman2016elbo}, \citet{chenisolating} refined the decomposition and separated TC of $\rvz$ from other terms. Since the independence of latent variables can be one of the sources of disentanglement, they introduced $\beta$-TCVAE with a new objective function that penalizes TC in order to learn factorized representation. At the same time, also recognizing the importance of TC in factorizing aggregated posterior, \citet{kim2018disentangling} independently introduced FactorVAE that penalizes $TC(\rvz)$ with a different implementation. Such strategy of penalizing $TC(\rvz)$ can be formulated as
\begin{align}
    \mathcal{L}_{\beta-\mathrm{TC}} &:=  \mathrm{ELBO}  - \beta \mathrm{TC}(z).
\end{align}
Though this strategy are effective at factorizing aggregated posterior \cite{locatello2018challenging}, according to Theorem \ref{tcbound}
the mean representations can still be entangled. 
To resolve this, we propose a regularized TC-VAE (RTC-VAE),
\begin{align}\label{rtcVAEobj}
    \mathcal{L}_{\mathrm{RTC}} := \mathcal{L}_{\beta-\mathrm{TC}} - \eta \cdot \text{tr}(\E_{p(n)} Cov_{q(z|n)}[z]),
\end{align}
where
$
     \text{tr}(\E_{p(n)} Cov_{q(z|n)}[z]) = \sum_k^D \E_{p(n)}[\sigma_k^2(n)].
$
Our penalty originates from the first term of the law of total covariance: $$Cov_{q(z)}[z] = \E_{p(n)} Cov_{q(z|n)}[z] + Cov_{p(n)}(\E_{q(z|n)}[z]).$$ 
Note that a factorized distribution 
$q(z)$ must have a diagonal covariance matrix $Cov_{q(z)}[z]$. 
For VAEs, it is conventional to set $q(z|n)$ as a factorized distribution, e.g. $\mathcal{N}(\mu(n), \text{diag} \sigma^2(n))$, which forces VAEs to behave similar to PCA (\citet{rolinek2019variational}). As a result, the first term $\E_{p(n)} Cov_{q(z|n)}[z]$ is forced to be diagonal.

Motivated by this, \citet{kumar2017variational} proposed DIP-VAEs which penalizes the off-diagonal terms in the second term in the law of total correlation and ignores the first term in order to get a diagonal covariance matrix $Cov_{q(z)}[z]$. 
\citet{locatello2018challenging} recognized DIP-VAEs being effective on factorizing aggregated posterior, but we point out that this is actually mistaken. The reason is simply because zero correlation does not necessarily imply independence (see details in Section~\ref{Experiments}).

Our approach, on the other hand, does not penalize directly on $\vmu$. Instead, we penalize on $\evsigma$, the standard deviation of the distribution $q(z|n)$, which is the first term in the law of total covariance. This may seem little counter-intuitive at first sight, since penalizing a diagonal component of covariance $Cov[\rvz]=Cov_{q(z)}[z]$ seems not helpful to factorising. However, in the view of Theorem \ref{tcbound}, the additional regularizer in \eqref{rtcVAEobj} will force the distribution of $\rvz$ to be similar to the distribution of $\vmu$. Hence, it pushes us away from the situation of large $\mathrm{TC}(\vmu)$ and low $\mathrm{TC}(\rvz)$. Consequently, by minimizing $\mathrm{TC}(\rvz)$ we get low $\mathrm{TC}(\vmu)$, which leads to factorized mean representations.


In practice, we keep $\text{tr}(\E_{p(n)} Cov_{q(z|n)}[z])$ in a range, e.g., $(0.01, 0.04)$, by multiplying the hyperparameter $\eta$ with 1.2 if this term is greater than $0.04$ and set $\eta$ to zero if the term is smaller than $0.01$. If the variance of latent variables vanishes completely, the VAE degenerates to a deterministic autoencoder. Thus, this approach saves the effort of tuning extra hyperparameter and keeps the stochastic nature of VAE. 

\section{Estimation of Total Correlation}
\label{sec:TCestimate}

\begin{figure*}[t]
\centering
    \includegraphics[width=\linewidth]{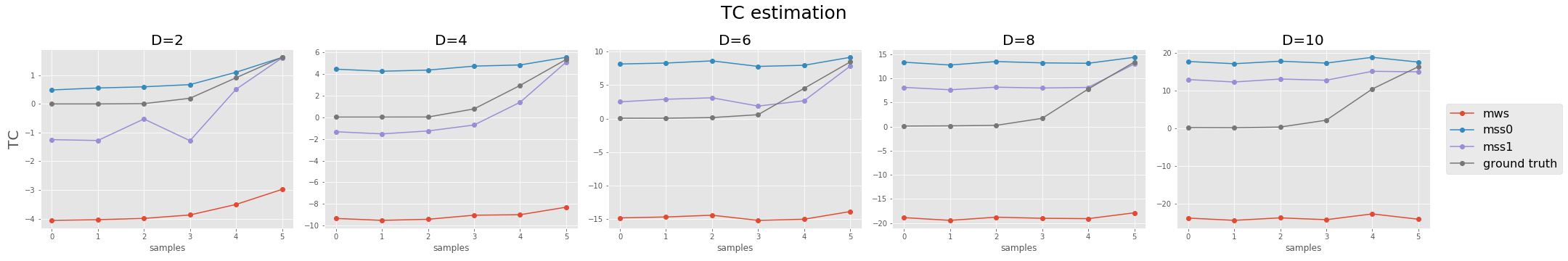}
    \caption{Let $\vmu \sim \mathcal{N}(0,\mathbf{\mSigma})$, and $\rvz|\mu \sim \mathcal{N}(\mu, \mSigma')$ where $\mSigma' = \mathrm{diag}(\sigma^2)$ and $\sigma=0.1$. The x-axis of each plot shows the determinant of $\mathbf{\mSigma}$ changes from 0 to 1, and y-axis shows $\mathrm{TC}(\rvz)$. Compared with ground truth $\mathrm{TC}(\rvz)$ (calculated by Proposition~\ref{tc_normal}), higher dimension will cause larger error in TC estimation, especially when TC is low.}
    \label{fig:estimators_gt2}
\end{figure*}
To calculate the objective function \eqref{rtcVAEobj}, a key step is to estimate TC. For multivariate normal distribution\footnote{
One may choose other prior distributions for a VAE model for different reasons. Here, normal distribution helps our analysis and simplifies the scenario. This is the reason why we choose normal distribution as prior.
}, 
its ground truth TC can be explicitly calculated thanks to the following proposition,
\begin{proposition}\label{tc_normal}
Let $\rvx \sim \mathcal{N}(0, \mSigma)$, then
\begin{align}
    \mathrm{TC}(\rvx) = \frac 1 2 \left( \mathrm{log} |\mathrm{diag} (\mSigma)| - \mathrm{log} |\mSigma|\right).
\end{align}
\end{proposition}
Proposition \ref{tc_normal} is a simple result, which is why its exact originality is difficult to track, but it is quite handy in our analysis. In Appendix~\ref{app:GaussianTC}, we provide a simple proof for the convenience of readers. \citet{locatello2018challenging} even used this proposition to approximate the TC of the mean representations in latent space. 

To estimate TC, naive Monte Carlo method comes with an intrinsic issue of underestimation. To resolve this, \citet{kim2018disentangling} proposed a discriminator network with the help of \textit{density-ratio trick} (see equation (3) and Appendix D. of \citet{kim2018disentangling}). In \citet{chenisolating}, two kinds of estimator of TC are proposed, Minibatch Weighted Sampling (MWS) and Minibatch Stratified Sampling (MSS) (see definitions in Appendix \ref{app:TC_estimators_compare}). 

In this work, we adopt density-ratio trick as our main method for estimating the total correlation of RTC-VAE. The reason is that we found out there exist some problems of MWS and MSS: the curse of dimensionality and an unintended latent dimension shutdown.

Our anaysis on MWS and MSS consists of both experimental and theoretic analysis.
First, we evaluate MWS, MSS$_0$ and MSS$_1$ (see definitions in Appendix \ref{app:TC_estimators_compare}) through the following experiments. 
Let $\vmu \sim \mathcal{N}(0,\mSigma)$ where $\mathrm{diag}\mSigma=\mathbf{I}$, and $\rvz|\mu \sim \mathcal{N}(\mu, \mSigma')$ where $\mSigma' = \mathrm{diag}(\sigma^2)$ and $\sigma=0.1$. We set $\sigma$ small so that the distribution of $\rvz$ can be approximated by normal distribution, and the ground truth $\mathrm{TC}(\rvz)$ can be calculated by Proposition~\ref{tc_normal}. Then by adjusting $|\mSigma|$, we can control $\mathrm{TC}(\rvz)$. We evaluate different estimators on different TC's, and results are presented in 
Figure~\ref{fig:estimators_gt2}.

From the experiments, we summarize some observations: 1. MWS tends to underestimate TC in general; 2. For latent space of dimension $<4$, MSS$_0$ and MSS$_1$ are relatively accurate; 3. For latent space of high dimension, both MSS$_0$ and MSS$_1$ tend to overestimate TC when the actual value of TC is small; 4. Overall MSS$_1$ estimates closer to ground truth than MMS$_0$ does.

In Appendix~\ref{app:TC_estimators_compare}, we provides a theoretic analysis of the 3rd observation to explain why these estimators deteriorate as dimension increases. In addition, we find that MWS and MSS may lead to an unintended shutdown of latent dimensions. While shutting down dimensions may not necessarily hurt disentanglement (it even can be helpful), the shutdown caused by these estimators is yet less understood for practice. We refer readers to Appendix~\ref{app:TC_estimators_compare} for detailed analysis.




\textbf{Density-ratio trick and auxiliary discriminator: }The gist of \textit{density-ratio trick} is to estimate the KL-divergence between the distribution of the latent representation $q(z)$ and the distribution of factorized latent representations $\prod_j q(z_j)$, which can be described as following \citep{nguyen2010estimating,sugiyama2012density},
\begin{align}\label{densityRatioTrick}
    \mathrm{TC}(\rvz) 
            \approx \E_{q(z)}\left[\log\frac{D(z)}{1-D(z)} \right],
\end{align}
where $D$ is discriminator that classifies $\rvz$ being sampled from $q(z)$ or $\prod_j q(z_j)$. 
We implemented our TC estimator according to Section 3 in \citet{kim2018disentangling} for training the auxiliary network $D$.

We do not include a direct numerical comparison between density-ratio trick and MWS/MSS because density-ratio trick would gain unfair advantage due to a potentially overfitting auxiliary discriminator.

\section{Experiments}
\label{Experiments}

We compare RTC-VAE with three models: FactorVAE and DIP-VAE-I\&II. We saved the experiments on $\beta$-TCVAE because its problem of disparity is the same as FactorVAE \citep{locatello2018challenging}. 
The datasets we use include dSprites~\citep{dsprites17}, Shapes3D~\citep{3dshapes18} and Car3d~\citep{reed2015deep}.
Check Table~\ref{architecture} for the architectures of encoder and decoder and Table~\ref{hyperparameters} for hyper-parameter setting. More details of experiments can be found in Appendix~\ref{app:experimentdetails}. 
We use the same structure for discriminator as \citet{kim2018disentangling} suggested for FactorVAE, which is a 6-layer MLP with 1000 hidden units per layer and leaky ReLU activation.

For RTC-VAE, we set the hyperparameter $\eta=\max(10,\beta)$. We bound $\eta$ from below to avoid the situation where the variance term in \eqref{rtcVAEobj} is so small that it will not contribute much compared with the $\mathcal{L}_{\beta-TC}$ term. Especially when $\beta$ is small, we need $\eta$ strong enough to regularize $TC_{sample}$. 

We choose batch size 500 for all models on all data sets to balance between performance and training time, whereas \citet{locatello2018challenging,kim2018disentangling} used 64, \citet{kumar2017variational} used 400 and \citet{chenisolating} used 2048 to account for the bias in minibatch estimation. Learning rate is fixed to $1\times10^{-3}$.
We evaluate models 
on 5000 randomly sampled data on every data set.

We estimate $TC_{mean}$ and $TC_{sample}$ by Proposition~\ref{tc_normal} as proposed by \citet{locatello2018challenging}. Specifically, we calculate the correlation matrices of the mean and sampled latent vectors, $\mu$'s and $z$'s, encoded from the 5000 samples. 


\subsection{Eliminating the Disparity between $TC_{sample}$ and $TC_{mean}$}

We first show that RTC-VAE has eliminates the disparity between $TC_{sample}$ and $TC_{mean}$. Again, since $\beta$-TCVAE has the same problem of disparity as FactorVAE, it is sufficient to compare RTC-VAE with FactorVAE. In order to do that, we evaluate $TC_{sample}$ and $TC_{mean}$ of RTC-VAE under different regularization strength and compare them with the corresponding values of FactorVAE. 

\begin{figure}[t]
\centering



\hbox{\hspace{-0.9cm}\includegraphics[width=1.25\columnwidth]{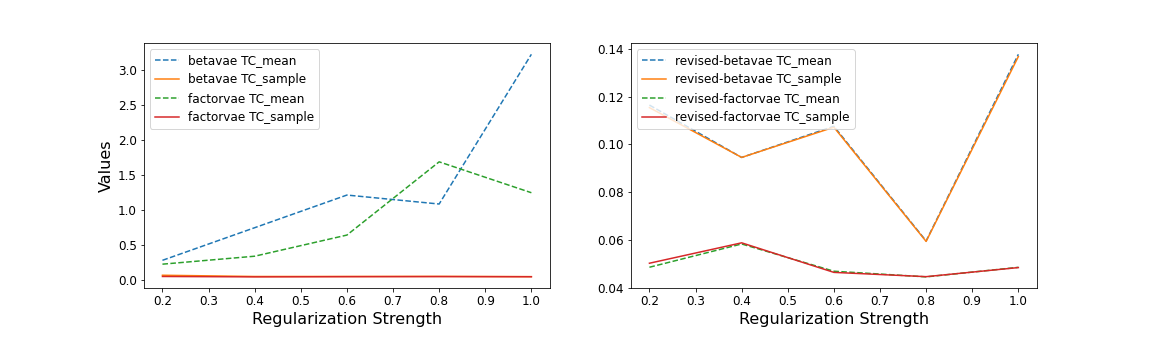}}



\caption{
Results on dSprites under all regularization strength: (Left) The disparity exists between dashed lines $TC_{mean}$ and solid lines $TC_{sample}$ for both $\beta$-VAE and FactorVAE. (Right) For revised models, the disparity is eliminated.}
\label{fig:TC_disparity}
\end{figure}

\begin{figure}[t]
\centering
\hbox{\hspace{-0.8cm}\includegraphics[width=1.2\columnwidth]{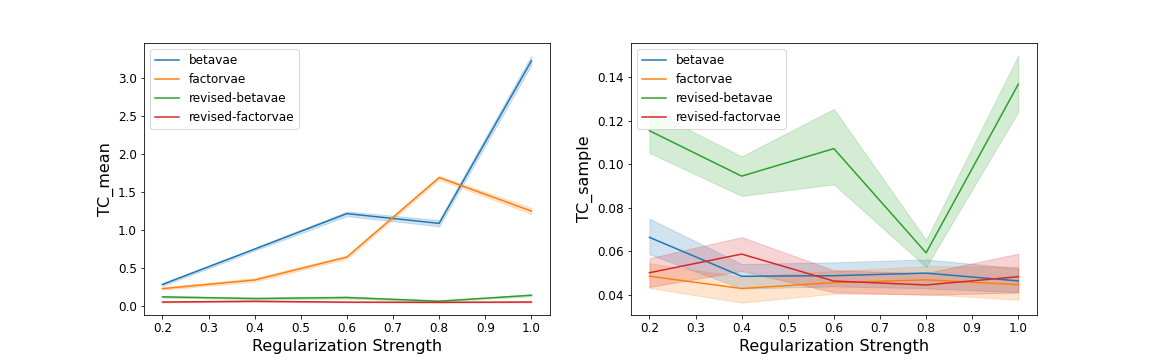}}
\caption{Results on dSprites: (Left) $TC_{mean}$ of revised models are much lower than vanilla models. (Right) $TC_{sample}$ of revised models are comparable to vanilla ones.
}
\label{fig:TC_disparity_a}
\end{figure}

In Figure~\ref{fig:TC_disparity}, we see that:
(Left) Under all regularization strength, the disparity exists between $TC_{mean}$ and $TC_{sample}$ for both $\beta$-VAE and FactorVAE. (Right) With the regularizer on variance of $q(z|n)$, there is almost no difference between $TC_{sample}$ and $TC_{mean}$, i.e., the disparity is evidently remedied.
In Figure \ref{fig:TC_disparity_a}, the revised $\beta$-VAE and FactorVAE can obtain much lower $TC_{mean}$ and than vanilla ones, meanwhile their $TC_{sample}$ are comparably low.

\subsection{Factorizing Aggregated Posterior}\label{factor_agg_post}

\citet{locatello2018challenging} reported that DIP-VAEs seem to be immune to the disparity between TCs and pointed out it is due to the measurement of TC being Gaussian based. Since DIP-VAEs are trained by regularizing the off-diagonal elements of correlation matrix, by Proposition~\ref{tc_normal} they are guaranteed to have low estimated TCs. Then the question is: Do DIP-VAEs obtain really factorized representation? Our investigation shows that the answer is no. 

\begin{figure}[t]
\begin{subfigure}[h]{0.49\columnwidth}
\raisebox{-\height}{\includegraphics[width=\columnwidth]{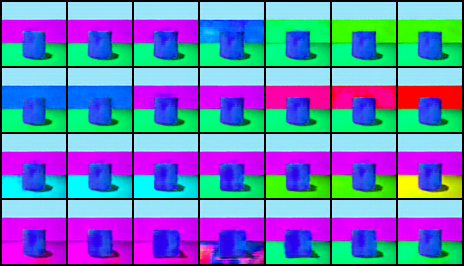}}
\centering
\caption{DIP-VAE-I latent walk $z_1,z_3,z_0,z_9$}
\end{subfigure}
\hfill
\begin{subfigure}[h]{0.49\columnwidth}
\raisebox{-\height}{\includegraphics[width=\columnwidth]{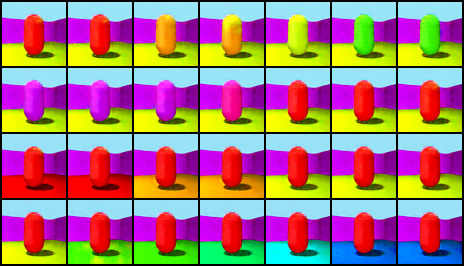}}
\centering
\caption{RTC-VAE latent walk  $z_1,z_5,z_0,z_7$}
\end{subfigure}
\begin{subfigure}[h]{0.49\columnwidth}
\raisebox{-\height}{\includegraphics[width=\columnwidth]{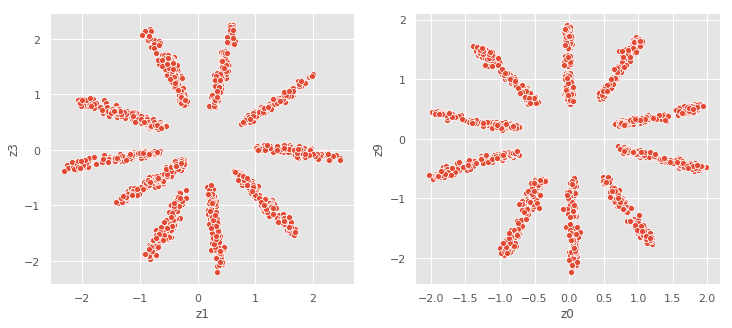}}
\centering
\caption{DIP-VAE-I pairplot $z_1\&z_3$, $z_0\&z_9$}
\end{subfigure}
\hfill
\begin{subfigure}[h]{0.49\columnwidth}
\raisebox{-\height}{\includegraphics[width=\columnwidth]{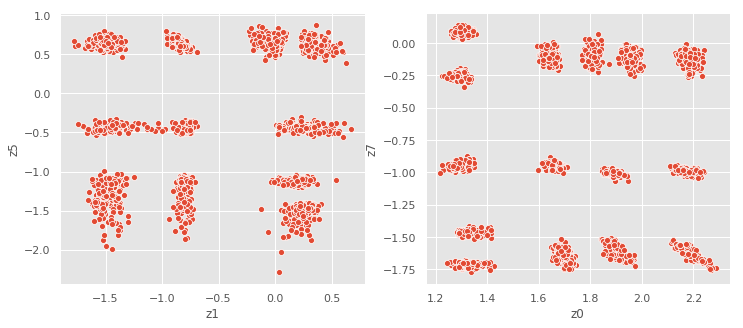}}
\centering
\caption{RTC-VAE pairplot $z_1\&z_5$, $z_0\&z_7$}
\end{subfigure}
\caption{Each row is a latent walk of one dimension of a VAE in (a) and (b). (a) In a DIP-VAE-I model, both $z_1\&z_3$ represent wall hue, and both $z_0\&z_9$ represent floor hue. Notice that each dimension covers different subsets of wall hue or floor hue, even though they are almost uncorrelated (see (c)). (b) In an RTC-VAE model, both $z_1\&z_5$ represent object hue, and both $z_0\&z_7$ represent floor hue, though these dimensions are almost independent (see (d)). (c) Pairplots of $z_1\&z_3$, $z_0\&z_9$ of DIP-VAE show that their correlation is very low (due to its radial symmetry), but they are apparently not independent. (d) The pairplots of $z_1\&z_5$, $z_0\&z_7$ of RTC-VAE are very close to the behavior of discrete independent distributions. See full pairplots in Appendix~\ref{app:experimentdetails}.}
\label{fig:factor_disint}
\end{figure}

\begin{figure}[t]

\begin{subfigure}[h]{0.49\columnwidth}
\includegraphics[width=\linewidth]{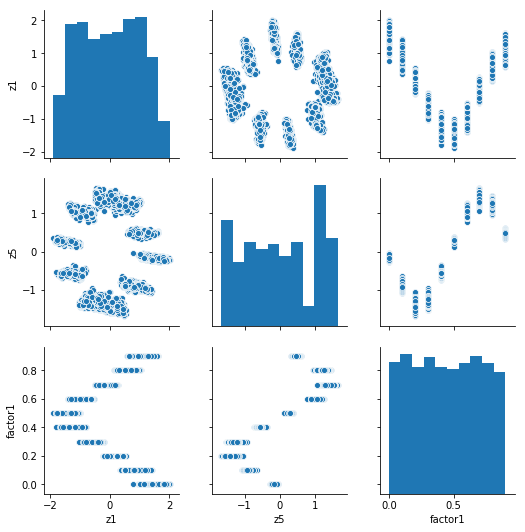}
\end{subfigure}
\hfill
\begin{subfigure}[h]{0.49\columnwidth}
\includegraphics[width=\linewidth]{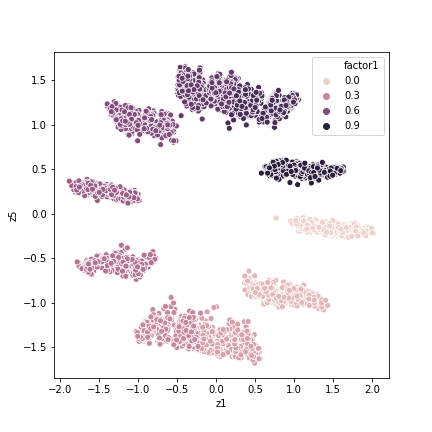}
\end{subfigure}
\caption{DIP-VAE-I on Shape3D. 
(Left) Though $z_1$ is almost uncorrelated to $z_5$, they are not independent. Meanwhile, both $z_1$ and $z_5$ are clearly linked to  factor 1.  (Right) A closer look, $z_1$ and $z_5$ together represent factor 1, uncorrelated but not independent.}
\label{fig:dip_factor_disint}
\end{figure}

In Figure~\ref{fig:factor_disint} (c) and also Figure~\ref{fig:dip_factor_disint}, we see that the learned latent representations of DIP-VAEs are uncorrelated but not factorized. 
Notice that estimating TC by Proposition \ref{tc_normal} is only valid for the multivariate normal distribution. If the presumption is violated, any uncorrelated distribution will have zero TC by such estimation. Since DIP-VAEs penalize directly on the correlation of mean representations, it leads to uncorrelated distributions and low TC estimation. Yet an uncorrelated and non-Gaussian distribution is not necessarily independent or factorized. 

On the other hand, RTC-VAE can successfully factorize representations (both sample and mean). Since the mean and sampled representation are very close, we only need to examine one of them.
A typical distribution of latent variables learned by RTC-VAE is presented in the pairplot of all latent variables, see Figure~\ref{fig:factor_disint} (d) and also Figure~\ref{fig:RTC_s3d_pairplot} in Appendix~\ref{app:experimentdetails}. We observe that the distribution present features of discrete independent distributions. 



\subsection{Factor Disintegration: Is Factorized Representation Disentangled?}
\label{factorDisintegrate}

Now that we can obtain factorized representations, the next question is: is a factorized representation disentangled?
Here, we point out \textit{factor disintegration}, indicating multiple \textit{independent} latent variables simultaneously represent one single factor of variation, exists in factorized representation, which maybe an unwanted feature for disentanglement. To this end, future study of disentangled representation should consider this tradeoff between factor disintegration and factorized representations. 

First, we will describe what factor disintegration is.
For example, in Shape3D, the wall hue is a 1-d factor taking values between 0 and 1. It turns out that a VAE model 
can cause the 1-d factor to disintegrate into 2 or more latent variables. Then each latent variable controls a subset of wall hue. 
In this way, even though the VAE can have a highly factorized latent representation, it manages to
use multiple dimensions to represent the wall hue instead of one, hence a factor disintegration (see Figure~\ref{fig:factor_disint} (b) and (d)). 


Factor disintegration disobeys the notion of ``compactness'' introduced by \citet{eastwood2018framework}, where compactness indicates each factor associates only one or a few latent variables. So, factor disintegration is a subclass of non-compactness (additionally presuming independence). Though there is still 
disagreement on whether compactness should be a character of disentanglement, e.g., \citet{ridgeway2018learning}, factor disintegration can potentially lead to unnecessarily many latent variables associating to a single factor of variation.

\begin{figure}[h]
\begin{subfigure}[h]{0.49\columnwidth}
\raisebox{-\height}{\includegraphics[width=\columnwidth]{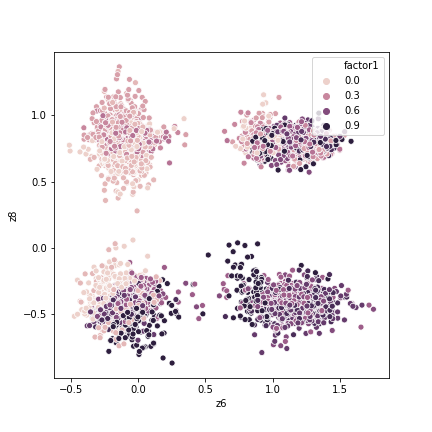}}
\centering
\end{subfigure}
\hfill
\begin{subfigure}[h]{0.49\columnwidth}
\raisebox{-\height}{\includegraphics[width=\columnwidth]{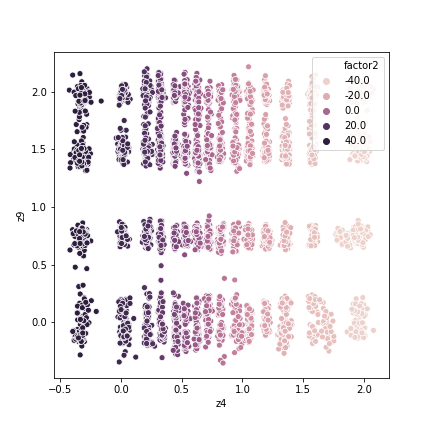}}
\centering
\end{subfigure}
\caption{A close look at the factorized latent distribution of RTC-VAE on Shapes3D. Each graph is a distribution of two latent variables, and the color labels how a factor of variation changes.
(Left) Affected by factor disintegration, Factor 1 (object hue) has top 2 $R^2$ scores with latent variables, $\rvz_6$ and $\rvz_8$, 0.23 and 0.17 respectively. Meanwhile, $\rvz_6$ and $\rvz_8$ appear to be independent, and their correlation coefficient $corr(\rvz_6, \rvz_8)=0.017$.
(Right) Factor 2 (orientation) is separately represented by $\rvz_4$, and the second top 2 correlated latent variable is $\rvz_9$, and the corresponding $R^2$ scores are 0.97 and $1\times10^{-4}$. $corr(\rvz_4, \rvz_9)=-0.01$. Then Factor 1 and 2 contribute 0.06 and 0.97 respectively to SAP score.}
\label{fig:factorized_entangled}
\end{figure}

\begin{figure}[t]
\begin{subfigure}{0.45\columnwidth}
\raisebox{-\height}{\includegraphics[width=\columnwidth]{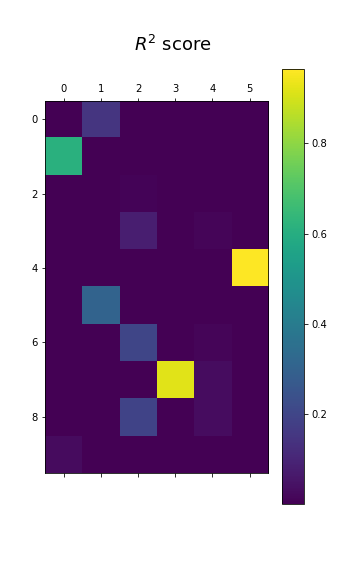}}
\end{subfigure}
\hfill
\begin{subfigure}{0.45\columnwidth}
\raisebox{-\height}{\includegraphics[width=\columnwidth]{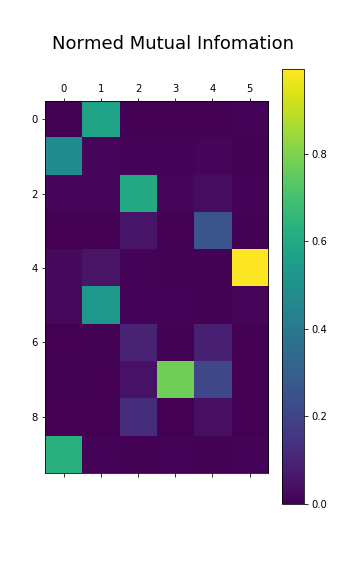}}
\end{subfigure}
\caption{Evaluating an RTC-VAE model with $R^2$ scores and normalized mutual information on Shape3D, which are used to calculate SAP and MIG respectively. Columns are factors of variation (ground truth), and rows are latent variables. When the gap between the top 2 scores in a column is small then the metric is suppressed. For example, for factor $1$, the highest and second highest scoring variables for both metrics are ($z_0$, $z_5$) and the gaps between them are small. (Right) Mutual information discovers more factor disintegration, e.g., factor 0$\&$4 have two high scores and the gaps are small (see Section~\ref{factorDisintegrate}). Hence, MIG also gets suppressed (see Section~\ref{metrics}). } 
\label{fig:R2_MI_s3d_rtc}
\end{figure}

\subsection{The Effect of Factor Disintegration on Disentanglement Metric}
\label{metrics}

To further demonstrate how factor disintegration will affect disentangled learning, we analyze its effect on disentanglement metric. 

There are many disentanglement metrics,
and most of these metrics share something in common. They look for the link between each factor of variation and each latent variable though the way of measuring the link differs.
For example, \citet{chenisolating} proposed \textit{mutual information gap} (\textbf{MIG}), which estimates the mutual information between each latent variable and each ground truth factor and then find the average gap between top 2 scores. \textbf{DCI} \citep{eastwood2018framework} computes the uncertainty (entropy) of predicting ground truth factor by latent variables, and then constructs a weighted average as a score. 
\citet{higgins2017beta} proposed \textbf{BetaVAE}, a linear classifier that predicts the index of a fixed factor of variation. Then they use the accuracy of the classifier as a disentanglement metric. \citet{kim2018disentangling} improved this method by using majority vote classifier to handle some corner case. 

As \citet{locatello2018challenging} pointed out, most metrics could actually be mildly correlated, and pairs (BetaVAE, FactorVAE) and (MIG, DCI) are even strongly correlated with each other. 
It implies that no matter which metric we use, if a model is affected by factor disintegration, it can find multiple latent variables scoring similar values, and hence suppress its final score under such metric.

In the following, we test the argument above with two disentanglement metrics:
MIG and \textit{Separated Attribute Predictability} score (\textbf{SAP score}), proposed by \citet{kumar2017variational} . 
Both metrics are classifier-free and essentially independent of the data. 
The idea behind SAP is similar to MIG but the underline measurement is $R^2$ score instead of mutual information. Specifically, SAP computes $R^2$ score between each latent variable and ground truth factor, and then calculate the difference between top 2 scores for each ground truth factor, and lastly take the average of these differences as a final score. 
Considering the theoretically optimal case (i.e., every ground truth factor is linearly correlated with exactly one latent variable and uncorrelated with all other variables), SAP score has an optimal value 1 (see evaluation of models with SAP in Appendix~\ref{app:experimentdetails}), whereas MIG is bounded above by the average entropy of each ground truth factor. 


In Figure~\ref{fig:factorized_entangled} and Figure~\ref{fig:R2_MI_s3d_rtc}, we see that when factor disintegration happens, the gap between top 2 correlation (with factor) is dramatically suppressed. 
(see more details in Appendix~\ref{app:experimentdetails})

\section{Conclusion}
In this work, we theoretically explain the relation between TC of sampled and mean distribution. We analyze the methods of estimating TC and point out some unnoticed problem. We demonstrate that RTC-VAE can eliminates the disparity between TC of the sampled and mean representations. Also, we compare RTC-VAE with DIP-VAEs and point out that DIP-VAEs can end up with uncorrelated yet dependent latent variables. Last, we find out a tradeoff between factorizing aggregated posterior and factor disintegration underlines disentangling representation. 







\bibliography{rtcvae}

\begin{thebibliography}{}

\bibitem[Alemi et~al., 2016]{alemi2016deep}
Alemi, A.~A., Fischer, I., Dillon, J.~V., and Murphy, K. (2016).
\newblock Deep variational information bottleneck.
\newblock {\em arXiv preprint arXiv:1612.00410}.

\bibitem[Bengio et~al., 2013]{bengio2013representation}
Bengio, Y., Courville, A., and Vincent, P. (2013).
\newblock Representation learning: A review and new perspectives.
\newblock {\em IEEE transactions on pattern analysis and machine intelligence},
  35(8):1798--1828.

\bibitem[Bengio et~al., 2007]{bengio2007scaling}
Bengio, Y., LeCun, Y., et~al. (2007).
\newblock Scaling learning algorithms towards ai.
\newblock {\em Large-scale kernel machines}, 34(5):1--41.

\bibitem[Bouchacourt et~al., 2018]{bouchacourt2018multi}
Bouchacourt, D., Tomioka, R., and Nowozin, S. (2018).
\newblock Multi-level variational autoencoder: Learning disentangled
  representations from grouped observations.
\newblock In {\em Thirty-Second AAAI Conference on Artificial Intelligence}.

\bibitem[Burgess and Kim, 2018]{3dshapes18}
Burgess, C. and Kim, H. (2018).
\newblock 3d shapes dataset.
\newblock https://github.com/deepmind/3dshapes-dataset/.

\bibitem[Chen et~al., 2018]{chenisolating}
Chen, R.~T., Li, X., Grosse, R., and Duvenaud, D. (2018).
\newblock Isolating sources of disentanglement in vaes.
\newblock {\em arXiv preprint arXiv:1802.04942}.

\bibitem[Eastwood and Williams, 2018]{eastwood2018framework}
Eastwood, C. and Williams, C.~K. (2018).
\newblock A framework for the quantitative evaluation of disentangled
  representations.

\bibitem[Goodfellow et~al., 2009]{goodfellow2009measuring}
Goodfellow, I., Lee, H., Le, Q.~V., Saxe, A., and Ng, A.~Y. (2009).
\newblock Measuring invariances in deep networks.
\newblock In {\em Advances in neural information processing systems}, pages
  646--654.

\bibitem[Higgins et~al., 2017]{higgins2017beta}
Higgins, I., Matthey, L., Pal, A., Burgess, C., Glorot, X., Botvinick, M.,
  Mohamed, S., and Lerchner, A. (2017).
\newblock beta-vae: Learning basic visual concepts with a constrained
  variational framework.
\newblock {\em ICLR}, 2(5):6.

\bibitem[Hoffman and Johnson, 2016]{hoffman2016elbo}
Hoffman, M.~D. and Johnson, M.~J. (2016).
\newblock Elbo surgery: yet another way to carve up the variational evidence
  lower bound.

\bibitem[Kim and Mnih, 2018]{kim2018disentangling}
Kim, H. and Mnih, A. (2018).
\newblock Disentangling by factorising.
\newblock {\em arXiv preprint arXiv:1802.05983}.

\bibitem[Kingma and Welling, 2013]{kingma2013auto}
Kingma, D.~P. and Welling, M. (2013).
\newblock Auto-encoding variational bayes.
\newblock {\em arXiv preprint arXiv:1312.6114}.

\bibitem[Kumar et~al., 2017]{kumar2017variational}
Kumar, A., Sattigeri, P., and Balakrishnan, A. (2017).
\newblock Variational inference of disentangled latent concepts from unlabeled
  observations.
\newblock {\em arXiv preprint arXiv:1711.00848}.

\bibitem[LeCun et~al., 2015]{lecun2015deep}
LeCun, Y., Bengio, Y., and Hinton, G. (2015).
\newblock Deep learning.
\newblock {\em nature}, 521(7553):436--444.

\bibitem[Locatello et~al., 2018]{locatello2018challenging}
Locatello, F., Bauer, S., Lucic, M., Gelly, S., Sch{\"o}lkopf, B., and Bachem,
  O. (2018).
\newblock Challenging common assumptions in the unsupervised learning of
  disentangled representations.
\newblock {\em arXiv preprint arXiv:1811.12359}.

\bibitem[Matthey et~al., 2017]{dsprites17}
Matthey, L., Higgins, I., Hassabis, D., and Lerchner, A. (2017).
\newblock dsprites: Disentanglement testing sprites dataset.
\newblock https://github.com/deepmind/dsprites-dataset/.

\bibitem[Nguyen et~al., 2010]{nguyen2010estimating}
Nguyen, X., Wainwright, M.~J., and Jordan, M.~I. (2010).
\newblock Estimating divergence functionals and the likelihood ratio by convex
  risk minimization.
\newblock {\em IEEE Transactions on Information Theory}, 56(11):5847--5861.

\bibitem[Peters et~al., 2017]{peters2017elements}
Peters, J., Janzing, D., and Sch{\"o}lkopf, B. (2017).
\newblock {\em Elements of causal inference: foundations and learning
  algorithms}.
\newblock MIT press.

\bibitem[Reed et~al., 2015]{reed2015deep}
Reed, S.~E., Zhang, Y., Zhang, Y., and Lee, H. (2015).
\newblock Deep visual analogy-making.
\newblock In {\em Advances in neural information processing systems}, pages
  1252--1260.

\bibitem[Ridgeway and Mozer, 2018]{ridgeway2018learning}
Ridgeway, K. and Mozer, M.~C. (2018).
\newblock Learning deep disentangled embeddings with the f-statistic loss.
\newblock In {\em Advances in Neural Information Processing Systems}, pages
  185--194.

\bibitem[Rolinek et~al., 2019]{rolinek2019variational}
Rolinek, M., Zietlow, D., and Martius, G. (2019).
\newblock Variational autoencoders pursue pca directions (by accident).
\newblock In {\em Proceedings of the IEEE Conference on Computer Vision and
  Pattern Recognition}, pages 12406--12415.

\bibitem[Sugiyama et~al., 2012]{sugiyama2012density}
Sugiyama, M., Suzuki, T., and Kanamori, T. (2012).
\newblock Density-ratio matching under the bregman divergence: a unified
  framework of density-ratio estimation.
\newblock {\em Annals of the Institute of Statistical Mathematics},
  64(5):1009--1044.

\bibitem[Thomas et~al., 2018]{thomas2018disentangling}
Thomas, V., Bengio, E., Fedus, W., Pondard, J., Beaudoin, P., Larochelle, H.,
  Pineau, J., Precup, D., and Bengio, Y. (2018).
\newblock Disentangling the independently controllable factors of variation by
  interacting with the world.
\newblock {\em arXiv preprint arXiv:1802.09484}.

\end{thebibliography}
\bibliographystyle{apalike}

\newpage
\onecolumn
\section*{Appendix}
\setcounter{section}{0}

\section{Problems in Methods of Minibatch Estimators}
\label{app:TC_estimators_compare}

\subsection{Minibatch Weighted Sampling (MWS)}
\label{MWS}
See \citet{chenisolating},
\begin{align}\label{eq:MWS}
    \E_{q(z)}[\log q(z)] \approx \frac{1}{M}\sum_{i=1}^M\left[\log\sum_{j=1}^M q(z(n_i)|n_j)-\log(NM)\right]
\end{align}


\subsection{Minibatch Estimators (MSS)}
\label{app:MSS}
MSS can be described as:
For a minibatch of sample, $B_{M+1} = \{n_1,\ldots,n_{M+1}\}$,
\begin{align}\label{eq:MSS}
 \E_{q(z,n)}[\log q(z)] \approx \frac{1}{M+1} &\sum_{i=1}^{M+1} \log f(z_i, n_i, B_{M+1}\setminus\{n_i\}), 
\end{align}
where
\begin{align}
f(z, n^*, B_{M+1}\setminus\{n^*\}) & =  \frac{1}{N}q(z|n^*)\\
      + \frac{1}{M} \sum_{m=1}^{M-1} &q(z|n_m)  + \frac{N-M}{NM}q(z|n_m).
\end{align}
$f(z, n^*, B_{M+1}\setminus\{n^*\})$ is an unbiased estimator of $q(z)$, but it turns out when it is used for estimating TC, it suffers from the curse of dimensions.

\textbf{MMS$_0$ and MMS$_1$: } There is a small part of the implementation of MSS in Chen et al.'s code that is not quite clear to us, specifically, the computation of log importance weight matrix in \eqref{eq:MSS}. In our experiment, we implement MSS with our understanding and denote it as MSS$_1$, and we denote Chen et al.'s implementation MSS$_0$. The only difference is that we replace this chunk of code (\url{https://github.com/rtqichen/beta-tcvae/blob/master/vae_quant.py#L199-L201}) to
\begin{lstlisting}
for i in range(batch_size):
	W[i,i] = 1/N
	W[i,(1+i)%batch_size] = strat_weight
\end{lstlisting}

\subsection{Comparison of the Two Methods}

In Section \ref{sec:TCestimate} we present our empirical evaluation of MWS, MSS$_0$ and MSS$_1$, and the result is shown in Figure~\ref{fig:estimators_gt2}. Here, we conduct some theoretical analysis and try to explain some of our observations:
For latent space of dimension $<4$, MSS$_0$ and MSS$_1$ are relatively accurate; for latent space of high dimension, both MSS$_0$ and MSS$_1$ tend to overestimate TC when the actual value of TC is small. 

Recall the experiment settings: 
Let $\vmu \sim \mathcal{N}(0,\mSigma)$ where $\mathrm{diag}\mSigma=\mathbf{I}$, and $\rvz|\mu \sim \mathcal{N}(\mu, \mSigma')$ where $\mSigma' = \mathrm{diag}(\sigma^2)$ and $\sigma=0.1$. We set $\sigma$ small so that the distribution of $\rvz$ can be approximated by normal distribution, and the ground truth $\mathrm{TC}(\rvz)$ can be calculated by Proposition~\ref{tc_normal}. Then by adjusting $|\mSigma|$, we can control $\mathrm{TC}(\rvz)$. 


\subsubsection{The Curse of Dimensionality}

Let $M$ be the batchsize and $D$ be the dimensions of latent space.
Notice that $TC(\vmu)=0$ in the above setting and $TC(\rvz)$ is small.
Let $TC_{\rvz}$ be the estimation of $TC(\rvz)$ with minibatch estimator, either \eqref{eq:MSS} or \eqref{eq:MWS}, and we find that for $D\geq 5$ approximately
\begin{align}\label{tcmsseq1}
    TC_{\rvz} \approx O((D-1)\log M).
\end{align}
See details of deduction in Appendix~\ref{tcmss1}. Thus, $TC_{\rvz}$ seriously overestimates the true $TC(\rvz)$. 

\subsubsection{Unintended Shutdown of Latent Dimensions}


We also find that the estimation of TC by MSS and MWS is lower for distributions with few active latent dimensions (we refer it as dimension ``shutdown'') 
than distributions with fully active dimensions. To see why, consider $\vmu_{0}\sim \mathcal{N}(0, \sigma_0)$, where $\sigma_0\ll 1$, and $\vmu_{0-}\sim \mathcal{N}(0, \mathbf{Id}_{D-1})$ where $0-$ means all the dimensions except 0, and assume that $\vmu_0$ is uncorrelated with the rest, and $\rvz|\mu\sim \mathcal{N}(0, \mSigma)$, where $\mSigma=\mathrm{diag}(\sigma^2)$. Again, $TC(\vmu)=0$ and if we choose small $\sigma$, $TC(\rvz)$ is small.

Though a similar analysis we find that the estimation of $TC(\rvz)$ is approximately
\begin{align}\label{tcmsseq2}
    TC_{\rvz} \approx O((D-2)\log M).
\end{align}
See a proof in Appendix~\ref{tcmss2}.
Compared to \eqref{tcmsseq1}, the distribution with a shutdown dimension has a lower estimation of TC.
Hence, by penalizing estimated $TC_{\rvz}$ a model may converge to distribution with fewer active latent variables. 


We note that shutting down latent dimensions may be helpful to learning disentangled representation, 
e.g., if the number of ground truth dimensions can efficiently represent data, more dimensions may cause entanglement. There are some works studying the phenomenon of dimension shutdown of VAE, and readers may refer to \citet{rolinek2019variational} and reference therein. 
However, in the case of MWS/MSS, the shutdown may be unintended and it is yet unclear exactly how many dimensions get shutdown. In our opinion, in order to precisely induce dimension shutdown, a better solution may be introducing proper bias to models, which also motivates supervised learning to disentangled learning (in addition to regularizing factor disintegration discussed in Section~\ref{factorDisintegrate}).


\section{Proof of Theorem \ref{tcbound}}
\label{app:tcbound_proof}
In the following proof, we follow a convention of mathematical analysis: the meaning of $C$ can change through lines. Specifically, if there are $C_1$ and $C_2$, take $C=\max(C_1, C_2)$. Since we only care about boundedness of some quantity, this notation eliminates some redundant work of tracking. $B_R(c)=\{x\in\mathbb{R}^n : |x-c|<R\}$.

\begin{theorem*}[Theorem \ref{tcbound} restated]
Let $\vmu \sim \mathcal{N}(0, \mSigma)$. 
For a fixed $\mu$, let $\rvz|\mu \sim \mathcal{N}(\mu, \mSigma'(\mu))$, where $\mSigma'(\mu)$ is diagonal and satisfies that,
\begin{align}
    c_1\leq \sigma'_j(\mu) \leq c_2,
\end{align}
where $c_1,c_2>0$ and $\sigma'_j(\mu)$ is the diagonal element of $\mSigma'(\mu)$.
Then $TC(\rvz)$ is independent of $TC(\vmu)$ and 
\begin{align}
    \mathrm{TC}(\rvz) \leq C\frac{c_3^D}{c_1^D} \log \left(\frac{c_2}{c_1} \right) 
            + C\frac{c_3^{D+2}}{c_1^{D+2}},
\end{align}
where $c_3=\max(c_2, \sqrt{D})$ and $C$ is some constant that replies only on dimension $D$.
\end{theorem*}

Proof. Let \[S_+ = \{ z\in \R^D| p(z)\geq \prod_j p(z_j)\}, \quad
S_- = \{z\in \R^D| p(z)< \prod_j p(z_j)\},\] then
\begin{align*}
    \mathrm{TC}(\rvz) = \int_{S_+} + \int_{S_-} = \mathrm{TC}(\rvz)_+ + \mathrm{TC}(\rvz)_-.
\end{align*}
Since KL-divergence is non-negative, if $\mathrm{TC}(\rvz)_+$ is bounded, then $\mathrm{TC}(\rvz)$ must be bounded. In the following, we work on $S_+$, i.e., we assume $p(z)\geq \prod_j p(z_j)$.

Note that total correlation is invariant under scaling, i.e., for $\lambda>0$, $\lambda\vmu \sim \mathcal{N}(0, \lambda\mSigma)$, then $TC(\vmu)=TC(\lambda\vmu)$. In Gaussian case, one can see this by simply applying Proposition \ref{tc_normal}. Hence, let $\sigma_j$ be the standard deviation of $\vmu_j$, and we can assume $\max_j \sigma_j < 1$. Otherwise we can instead work on $\lambda\vmu$ with a sufficient small $\lambda$.

Fix some $R\geq1$, and for $|z|<R$,
\begin{align*}
    p(z) &= \E_{p(\mu)}[p(z|\mu)] = \int_{\mathbb{R}^D}p(\mu)p(z|\mu)d\mu  \\
        &= \frac{C}{c_1^{D}} \int_{\mathbb{R}^D} p(\mu) e^{-\frac{1}{2c_2^2}|z-\mu|^2} d\mu  \\
        &\leq \frac{C}{c_1^{D}} \int_{\mathbb{R}^D} p(\mu)  d\mu\\
        &= \frac{C}{c_1^{D}},
\end{align*}
And also for $|z| < R$,
\begin{align*}
    \prod_j p(z_j) &\geq \prod_j\int_{|\mu_j|<R} p(\mu_j) \frac{1}{\sqrt{2\pi c_2^2}} e^{-\frac{(z_j-\mu_j)^2}{2c_1^2} } d\mu_j \\
        &\geq \frac{C}{c_2^D}   \int_{|\mu|<R} e^{-\frac{|z-\mu|^2}{2c_1^2}}\prod_j p(\mu_j)  d\mu_1\cdots d\mu_D \\
        &\geq \frac{C}{c_2^D} e^{-\frac{2R^2}{c_1^2}} \frac{1}{2^D} \\
        &\geq \frac{C}{c_2^D} e^{-\frac{2R^2}{c_1^2}},
\end{align*}
where in the second last inequality we use the fact $|z-\mu|<2R$ and the fact $\max_j \sigma_j < 1$ and $R\geq1$.

Let $\Bar{\sigma}$ be the largest singular value of $\mSigma$, and for $|z|>R\geq 1$ we have, 
\begin{align*}
    p(z) &= \int_{B_{\frac{|z|}{2}}(0)} p(\mu)p(z|\mu)d\mu + \int_{B^c_{\frac{|z|}{2}}(0)}p(\mu)p(z|\mu)d\mu \\
    &\leq \int_{B_{\frac{|z|}{2}}(0)} p(\mu)\frac{C}{c_1^D}e^{-\frac{|z-\mu|^2}{2c^2_2}}d\mu 
        + \int_{B^c_{\frac{|z|}{2}}(0)} p(\mu)\frac{C}{c_1^D}e^{-\frac{|z-\mu|^2}{2c^2_2}}d\mu \\
    &\leq \frac{C}{c_1^D}e^{-\frac{||z|-\frac{|z|}{2}|^2}{2c^2_2}} \int_{B_{\frac{|z|}{2}}(0)} p(\mu) d\mu
        + \frac{C}{c_1^D} \int_{B^c_{\frac{|z|}{2}}(0)} p(\mu)d\mu \\
    &\leq  \frac{C}{c_1^D}e^{-\frac{|z|^2}{8c^2_2}} \cdot 1 
        + \frac{C}{c_1^D} \int_{|\mu|>\frac{|z|}{2}} p(\mu)d\mu \\
    &\leq \frac{C}{c_1^D}e^{-\frac{|z|^2}{8c^2_2}} 
        + \frac{C}{c_1^D} \int_{|\tau|>\frac{|z|}{2\Bar{\sigma}}} e^{-\frac{|\tau|^2}{2}} d\tau 
        \quad \text{(Let $\tau=T\mu$ where $T=(\mSigma^{-1})^{\frac{1}{2}})$} \\
    &\leq \frac{C}{c_1^D}e^{-\frac{|z|^2}{8c^2_2}} 
        + \frac{C}{c_1^D} \prod_j \int_{t>\frac{|z|}{2\Bar{\sigma}\sqrt{D}}} e^{-\frac{t^2}{2}} dt \\
    &\leq \frac{C}{c_1^D}e^{-\frac{|z|^2}{8c^2_2}} 
        + \frac{C}{c_1^D} e^{-\frac{|z|^2}{8D\Bar{\sigma}^2}} \left(\frac{|z|}{2\Bar{\sigma}\sqrt{D}}\right)^{-D} \\
    &\leq \frac{C}{c_1^D}e^{-\frac{|z|^2}{8c^2_3}},
\end{align*}
where $c_3 = \max(\sigma_2, \sqrt{D}\Bar{\sigma})$. The second last inequality is due to the estimation of complementary error function \cite{abramowitz1972handbook}. Again, we can scale $\lambda\vmu$ such that $\Bar{\sigma}$ of $\lambda\mSigma$ is less than 1. Hence, we can set $c_3 = \max(\sigma_2, \sqrt{D})$.

Also for $|z|>R$, 
\begin{align*}
    \prod_j p(z_j) &\geq \prod_j \left(\int_{|\mu_j|<R} p(\mu_j) \frac{C}{ c_2} e^{-\frac{|z_j-\mu_j|^2}{2c_1^2} } d\mu_j \right) \\
        &\geq \frac{C}{c_2^D}  \int_{|\mu_j|<\sigma_j, j=1\ldots D} e^{-\frac{|z-\mu|^2}{2c_1^2} }\prod_j  p(\mu_j)  d\mu_1\cdots d\mu_D  \\
        &\geq \frac{C}{c_2^D}  e^{-\frac{|2|z||^2}{2c_1^2} }\prod_j \left(\frac{1}{2}\right)  \\
        &\geq \frac{C}{c_2^D} e^{-\frac{2|z|^2}{c_1^2} }.
\end{align*}
Thus,
\begin{align*}
    \mathrm{TC}(\rvz) &= \E_{p(z)} \left[\log \frac{p(z)}{\prod_j p(z_j)} \right] \\
        &\leq \int_{B_R(0)}p(z)\log \frac{p(z)}{\prod_j p(z_j)} dz 
            + \int_{B^c_R(0)}p(z)\log \frac{p(z)}{\prod_j p(z_j)} dz \\
        &\leq \int_{B_R(0)} \frac{C}{c_1^D} \log \left(C\frac{c^D_2}{c_1^D} e^{\frac{2R^2}{c_1^2}}\right) dz
            + \int_{B^c_R(0)} \frac{C}{c_1^D} e^{-\frac{|z|^2}{8c_3^2}} \log C\frac{c^D_2}{c_1^D} e^{\frac{2|z|^2}{c_1^2} - \frac{|z|^2}{8c_3^2}} dz \\
        &\leq \frac{C}{c_1^D} \log \left(C\frac{c^D_2}{c_1^D} e^{\frac{2R^2}{c_1^2}}\right) R^D 
            + \int_{B^c_R(0)} \frac{C}{c_1^D} e^{-\frac{|z|^2}{8c_3^2}} \left(\log C\frac{c^D_2}{c_1^D} + \frac{2|z|^2}{c_1^2} \right) dz \\
        &\leq \frac{C}{c_1^D} \log \left(C\frac{c^D_2}{c_1^D} e^{\frac{2R^2}{c_1^2}}\right) R^D
            + \frac{C}{c_1^D} \log \left(C\frac{c^D_2}{c_1^D} \right) \prod_j \int_{|z_j|>\frac{R}{\sqrt{D}}} e^{-\frac{|z_j|^2}{8c_3^2}} dz_j
            + \frac{C}{c_1^{D+2}} \int_{|z|>R} e^{-\frac{|z|^2}{8c_3^2}}|z|^2 dz \\
        &\leq \frac{C}{c_1^D} \log \left(C\frac{c^D_2}{c_1^D} e^{\frac{2R^2}{c_1^2}}\right) R^D
            + C\frac{c_3^D}{c_1^D} \log \left(C\frac{c^D_2}{c_1^D} \right) \frac{e^{-\frac{R^2}{8c_3^2}D}}{(\frac{R}{\sqrt{8c_3^2D}})^D}
            + \frac{C}{c_1^{D+2}} c_3^2 e^{-\frac{R^2}{8c_3^2}} R^D \\
        &\leq C\frac{c_3^D}{c_1^D} \log \left(C\frac{c^D_2}{c_1^D} \right) 
            + C\frac{c_3^{D+2}}{c_1^{D+2}} \quad\text{(Take $R=c_3$ since it's arbitrary)} \\
        &\leq  C\frac{c_3^D}{c_1^D} \log \left(\frac{c_2}{c_1} \right) 
            + C\frac{c_3^{D+2}}{c_1^{D+2}}.
\end{align*}
To estimate the last term in the 4th inequality above, we transform the integral to spherical integral and then repeat integrate-by-part till we can estimate it with complementary error function.
$\Box$

One can directly check that the above argument is valid for $p(\mu)$ with compact support (then $c_3$ can be taken as $c_2$) or fast decay (faster than $O(e^{-|\mu|^2}$)). Specifically, if $p(\mu)$ has a compact support, the only thing will change is the estimation of $p(z)$ for $|z|>R\geq 1$. Scale $\vmu$ such that its support is contained in $B_1(0)$. Then the integral on $B^c_{\frac{|z|}{2}(0)}$ is 0 and hence $c_3$ can take $c_2$. If $p(\mu)$ decays faster than $O(e^{-|\mu|^2}$), then the argument is the same.

\section{Proof of Proposition \ref{tc_normal}}
\label{app:GaussianTC}

\begin{proposition*}[Proposition \ref{tc_normal} restated]
Let $\rvx \sim \mathcal{N}(0, \mSigma)$, then
\begin{align}
    \mathrm{TC}(\rvx) = \frac 1 2 \left( \mathrm{log} |\mathrm{diag} (\mSigma)| - \mathrm{log} |\mSigma|\right).
\end{align}
\end{proposition*}

Proof.
First, recall that the KL-divergence between two distributions $\Pdist$ and $\Qdist$ is defined as
\begin{align*}
\mathrm{KL}(\Pdist||\Qdist)= \E_{\Pdist}[\log\frac{\Pdist}{\Qdist}]
\end{align*} 
Also, the density function for a multivariate Gaussian distribution  $\mathcal{N}(\mu, \mSigma)$ is 
\begin{align*}
p(x) = \frac{1}{(2\pi)^{n/2} \mathrm{det}(\mSigma)^{1/2}}\mathrm{exp}(-\frac{1}{2}(x-\mu)^T\mSigma^{-1}
(x-\mu)). 
\end{align*}
Now, for two multivariate Gaussian $\Pdist_1$ and $\Pdist_2$, we have 
\begin{align*} 
\mathrm{KL}(\Pdist_1||\Pdist_2) &= \E_{\Pdist_1} [\mathrm{log} \Pdist_1- \log \Pdist_2] \\
        &=\frac{1}{2}\mathrm{log} \frac{\mathrm{det} \mSigma_2}{\mathrm{det} \mSigma_1} + 
         \frac{1}{2} \E_{p_1(x)} [-(x-\mu_1)^T\mSigma_1^{-1}(x-\mu_1)
        +(x-\mu_2)^T\mSigma_2^{-1}(x-\mu_2)] \\
        &= \frac{1}{2}\mathrm{log} \frac{\mathrm{det} \mSigma_2}{\mathrm{det} \mSigma_1}+ \frac{1}{2} \E_{p_1(x)}
        [-\mathrm{tr}(\mSigma_1^{-1}(x-\mu_1)(x-\mu_1)^T)+\mathrm{tr}(\mSigma_2^{-1}(x-\mu_2)(x-\mu_2)^T)] \\
        &= \frac{1}{2}\mathrm{log} \frac{\mathrm{det} \mSigma_2}{\mathrm{det} \mSigma_1}
        -\frac{1}{2}\mathrm{tr}(\mSigma_1^{-1}\mSigma_1)+\frac{1}{2} \E_{p_1(x)}[\mathrm{tr}(\mSigma_2^{-1}((xx^T-2x\mu_2^T+
        \mu_2\mu_2^T))] \\    
        &= \frac{1}{2}\mathrm{log} \frac{\mathrm{det} \mSigma_2}{\mathrm{det} \mSigma_1}
        -\frac{n}{2}+\frac{1}{2} \E_{p_1(x)}[\mathrm{tr}(\mSigma_2^{-1}((x-\mu_1)(x-\mu_1)^T+
        \underbrace{2(x-\mu_1)\mu_1}_{ \E_{p_1(x)}(x)= \mu_1}+\mu_1\mu_1^T
        -2x\mu_2^T+
        \mu_2\mu_2^T))] \\
        &= \frac{1}{2}\mathrm{log} \frac{\mathrm{det} \mSigma_2}{\mathrm{det} \mSigma_1}
        -\frac{1}{2}n + \frac{1}{2} \mathrm{tr}(\mSigma_2^{-1}(\mSigma_1+ (\mu_2-\mu_1)(\mu_2-\mu_1)^T)) \\
        &=\frac{1}{2}( \mathrm{log} \frac{\mathrm{det} \mSigma_2}{\mathrm{det} \mSigma_1}-n 
        +\mathrm{tr}(\mSigma_2^{-1}\mSigma_1) +(\mu_2-\mu_1)^T\mSigma_2^{-1}(\mu_2-\mu_1)) 
\end{align*}
            
Let $\Pdist$ be a multivariate Gaussian $\mathcal{N}(\mu, \mSigma_1)$, and then the product of the marginal distribution $\prod_i p_i(x)$ is also Gaussian $\mathcal{N}(\mu, \mSigma_2)$, where 
$\mSigma_2 = \mathrm{diag} (\mSigma_1)$.
Thus, the total correlation of multivariate Gaussian distribution is 
\begin{align*}
    \mathrm{TC}(\rvx) &= \mathrm{KL}(p(x)||\prod_i p_i(x)) \\
             &=\frac{1}{2}( \mathrm{log} \frac{\mathrm{det} \mSigma_2}{\mathrm{det} \mSigma_1} 
        -n + \mathrm{tr}(\mSigma_2^{-1}\mSigma_1) + (\mu-\mu)^T\mSigma_2^{-1}(\mu-\mu) )  \\
           &= \frac{1}{2}( \mathrm{log} \frac{\mathrm{det} \mSigma_2}{\mathrm{det} \mSigma_1} -n +n ) \\
           &=  \frac 1 2 \left( \mathrm{log} |\mathrm{diag} (\mSigma_1)| - \mathrm{log} |\mSigma_1|\right) 
\end{align*}
$\Box$

\section{Proof of \eqref{prob_gauss}}\label{app:prob_gauss_proof}
Proof. For $t>0$,
\begin{align*}
    P(|z^{(i)}-\mu^{(j)}|<t)
    & = P(|x|<t) \text{ where $x\sim \mathcal{N}(0,2)$}\\
    & = \int_{-t}^{t} \frac{1}{\sqrt{4\pi}} e^{-\frac{x^2}{4}} dx \\
    & = \sqrt{\int_{-t}^{t} \frac{1}{\sqrt{4\pi}} e^{-\frac{x^2}{4}} dx\int_{-t}^{t} \frac{1}{\sqrt{4\pi}} e^{-\frac{y^2}{4}} dy} \\
    & = \sqrt{\int_{-t}^{t}\int_{-t}^{t} \frac{1}{4\pi} e^{-\frac{x^2+y^2}{4}} dxdy} \\
    & = \sqrt{\int_{0}^{2\pi}\int_{0}^{t} \frac{1}{4\pi} e^{-\frac{r^2}{4}} rdr d\theta} \\
    & = \sqrt{1-e^{-\frac{t^2}{4}}} \\
    & = \frac{t}{2} + O(t^2) 
\end{align*}
$\Box$

\section{Sketched Proof of \eqref{tcmsseq1}}\label{tcmss1}

The following argument provides an approximated estimation of related quantities. The goal is not complete rigorousness but rather an intuitive yet quantitative explanation of our observations in Section \ref{sec:TCestimate}.

Recall that $TC(\vmu)=0$ and $\sigma=0.1$, and hence $TC(\rvz)$ is small. $M$ is batchsize and $D$ is latent dimension. Now,
consider $q(z_k^{(i)}|n^{(j)})$, where $(i,j,k)$ are indices of a box $(minibatch, minibatch, dimension)$ with size $M\times M\times D$ and let $n^{(j)}$ be a sample drawn in a minibatch and $z^{(i)}: = z(n^{(i)})$. We \textit{claim}: when the ground truth $TC(\rvz)$ is low,
considering $q(z_k^{(i)}|n^{(i)})$, only the elements on the diagonal plane of an index-box, namely those probabilities with indices $(i,i,k)$, take some bounded values $O(1)$, and all the other elements are very small. 

To rationalize our claim, it is obvious that $q(z_k^{(i)}|n^{(i)})$ is not small, and we only need to show the probability of $q(z_k^{(i)}|n^{(j)})$, $i\neq j$, being large is small enough to ignore for each minibatch. Let us first consider 1-D cases, where $\vmu \sim \mathcal{N}(0, 1)$, $\rvz|\mu\sim \mathcal{N}(0, \sigma^2)$. When $\sigma$ is small, $\rvz$ can be approximately treated as ${\mathcal{N}}(0, 1)$. $z^{(i)}$ and $\mu^{(j)}$ are independent for $i \neq j$, hence $z^{(i)}-\mu^{(j)} \sim \mathcal{N}(0, 2)$, and for any $t>0$, we can estimate the probability of $|z^{(i)}-\mu^{(j)}|<t$ by
\begin{align}\label{prob_gauss}
    P(|z^{(i)}-\mu^{(j)}|<t)
    = \frac{t}{2} + O(t^2) 
\end{align}
See a proof in Appendix \ref{app:prob_gauss_proof}.

Generalized to D-dimension, 
the probability $P(|z^{(i)}-\mu^{(j)}|<t)$ would be $\frac{t^D}{2^D} + O(t^{D+1})$\footnote{To see this, notice that the region within a hypersphere, $\{z^{(i)}: |z^{(i)}-\mu^{(j)}|<t\}$,  is contained in the hyper-rectangle, $\{z^{(i)}: |z^{(i)}_k-\mu^{(j)}_k|<t, k=1\ldots D\}$. Now, recall the assumption that TC$(\rvz)$ is small, implying the correlation among each components of $\rvz$ is low. Hence, the probability of the hyper-rectangle can be estimated simply by the product of the probability of each component.  
}.
Now, for the case $q(z_k^{(i)}|n^{(j)})$ being large, it happens only if $|z^{(i)}-\mu^{(j)}|<t$ and $t\leq 3\sigma$ (since the probability of normal distribution outside 3 standard deviation is very small). When $\sigma=0.1$, the probability of such cases to happen is $O(10^{-D})$. This means, when $i\neq j$, the number of such cases belongs to binomial distribution with $p=O(10^{-D})$ and $n$ less than batch-size. While batch-size usually is less than $10^3$, such cases can be ignored if $D\geq 5$ (both mean and variance are small). Therefore, we can assume $q(z_k^{(i)}|n^{(j)})$ is small for all $i\neq j$.

Thus,
\begin{align*}
    q(z_k^{(i)}|n^{(i)})= O(1), \quad q(z_k^{(i)}|n^{(j)})= o(1),
\end{align*}
and
\begin{align*}
    TC(\rvz)    
        & = \E_{q(z)}\left[\log\frac{q(z)}{\prod_k q(z_{k})}\right] \\
        & = \E_{q(z,n)}[\log q(z)] - \E_{q(z,n)}[\log \prod_k q(z_{k})] \\
        &\approx  \frac{1}{M}\sum_i\left(\log \frac{1}{M}\sum_{j} \prod_k q(z_k^{(i)}|n^{(j)}) - \log \prod_k \frac{1}{M}\sum_{j} q(z_k^{(i)}|n^{(j)})\right) \\
        &\approx  \frac{1}{M}\sum_i\left(\log \frac{1}{M}\sum_{j=i} O(1) - \log \prod_k \frac{1}{M}\sum_{j=i} O(1)\right) \\
        &\approx  \frac{1}{M}\sum_i\left(\log O(\frac{1}{M}) - \log  O(\frac{1}{M^D})\right) \\
        &\approx O((D-1)\log M).
\end{align*}
$\Box$

\section{Sketched Proof of \eqref{tcmsseq2}}\label{tcmss2}
Recall that the first dimension of $\vmu$ gets shutdown, i.e., $\vmu_{0}\sim \mathcal{N}(0, \sigma_0)$, where $\sigma_0\ll 1$, and $\vmu_{0-}\sim \mathcal{N}(0, \mathbf{Id}_{D-1})$. Then for any $t>\sigma_0$, $P(|z_0^{(i)}-\mu_0^{(j)}|<t)$ is O(1). For the rest of the dimensions, it reduces to $(D-1)$-dimension case (since true $TC(\rvz)$ is small, all dimensions can be treated independently). 
Hence,  $P(|z^{(i)}-\mu^{(j)}|<t)$ is approximately $\frac{t^{D-1}}{2^{D-1}} + O(t^{D})$.
Therefore, only probabilities with indices $(i,j,0)$ and $(i,i,k)$ where $k>0$ take some bounded values $O(1)$ and the rest can be ignored (for batchsize $M$, if $\sigma_0$ is sufficiently small, then we can choose $t$ such that $\frac{t^{D-1}}{2^{D-1}}\cdot M \ll 1$). 
Hence, $\frac{1}{M}\sum_{j} q(z_0^{(i)}|n^{(j)}) \approx \frac{1}{M}\sum_{j} O(1) \approx O(1)$, and
\begin{align*}
TC(\rvz)    
        & = \E_{q(z)}\left[\log\frac{q(z)}{\prod_k q(z_{k})}\right] \\
        &\approx \frac{1}{M}\sum_i\log\frac{\frac{1}{M}\sum_{j} \prod_k q(z_k^{(i)}|n^{(j)})}{\prod_k \frac{1}{M}\sum_{j} q(z_k^{(i)}|n^{(j)})} \\
        &\approx \frac{1}{M}\sum_i\log\frac{\frac{1}{M}\sum_{j}  (q(z_0^{(i)}|n^{(j)})\cdot \prod_{k>0} q(z_k^{(i)}|n^{(j)})) }{\prod_{k>0} \frac{1}{M}\sum_{j} q(z_k^{(i)}|n^{(j)})} \\
        &\approx \frac{1}{M}\sum_i\log\frac{\frac{1}{M} \cdot O(1)}{\prod_{k>0} \frac{1}{M} \cdot O(1)} \\
        &\approx \log O(M^{D-2}) \\
        &\approx O((D-2)\log M).
\end{align*}
$\Box$

The above argument can be easily generalized to the case of $S$-dimension shutdown till some integer $S\leq S_0\in(0,D)$. One reason for $S_0<D$ is that, the argument stops being true if $\frac{t^{D-S-1}}{2^{D-S-1}}\cdot M \ll 1$ no longer holds. After all, it is unlikely for a model to represent data with all latent dimensions shutdown. 

\section{Experiments}\label{app:experimentdetails}

\begin{table}[h]
\caption{Model's hyperparameters.} 
\begin{align*}
\begin{array}{ccc}
\hline
\textbf{Model} & \textbf{Parameter} & \textbf{Values} \\
\hline
\text{FactorVAE} & \gamma=\beta & [2,6,10,20,50] \\ 
\text{RTC-VAE} & \beta & [2,6,10,20,50] \\ 
\text{DIP-VAE-I} & \lambda_{od} & [2,5,10,20,50] \\ 
            & \lambda_d  & 10\lambda_{od}  \\
\text{DIP-VAE-II} & \lambda_{od} & [2,5,10,20,50] \\ 
            & \lambda_d  & \lambda_{od} \\
\hline
\end{array}
\end{align*}
\label{hyperparameters}
\end{table}

\begin{table*}[h]
\caption{Encoder and Decoder architecture. nc=number of channels}
    \begin{align*}
        \begin{array}{cc}
        \hline
        \textbf{Encoder} & \textbf{Decoder}  \\
        \hline
        \text{Input: nc} \times 64\times64  & \text{Input: } \R^{10} \\
        4\times4 \text{ conv, 32 ReLU, stride 2, padding 1} &  1\times1 \text{ upconv, 512 ReLU, stride 1} \\ 
        4\times4 \text{ conv, 64 ReLU, stride 2, padding 1} & 4\times4 \text{ upconv, 64 ReLU, stride 1} \\ 
        4\times4 \text{ conv,  64 ReLU, stride 2, padding 1}  & 4\times4 \text{ upconv, 64 ReLU, stride 2, padding 1}  \\
        4\times4 \text{ conv,  64 ReLU, stride 2, padding 1}  & 4\times4 \text{ upconv, 32 ReLU, stride 2, padding 1} \\
        4\times4 \text{ conv,  512 ReLU, stride 1}  & 4\times4 \text{ upconv, 32 ReLU, stride 2, padding 1}  \\
        1\times1 \text{ conv,  10, stride 1}  & 4\times4 \text{ upconv, nc ReLU, stride 2, padding 1}  \\
        \hline
        \end{array}
    \end{align*}
    \label{architecture}
\end{table*}

\begin{figure}[h]

\begin{subfigure}[h]{0.49\textwidth}
\includegraphics[width=\linewidth]{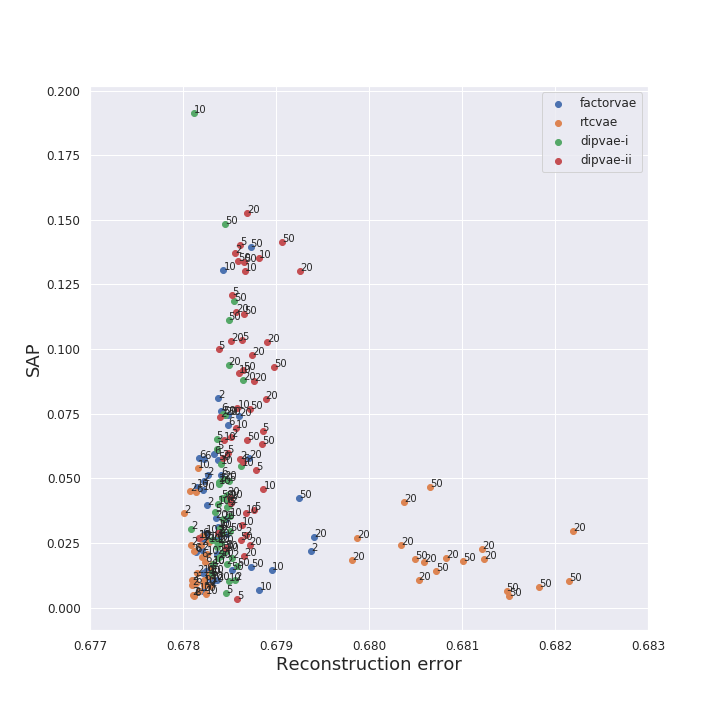}
\caption{The SAP score verses reconstruction error on dSprites.\\ 
For RTCVAE, $\beta>10$ can affect the quality of reconstruction.}
\end{subfigure}
\begin{subfigure}[h]{0.49\textwidth}
\includegraphics[width=\linewidth]{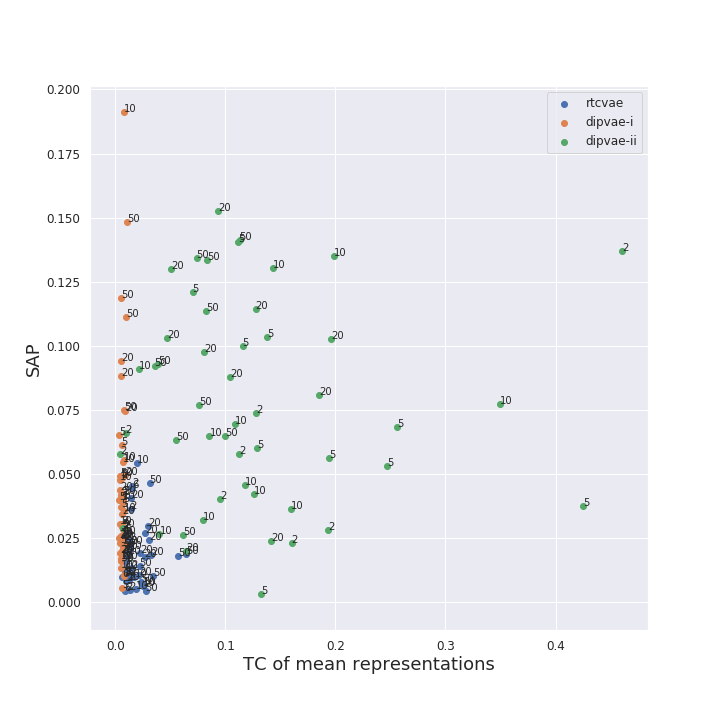}
\caption{The pair plot of $TC_{mean}$ and SAP shows no strong correlation, indicating that factorized aggregated posterior alone does not necessarily lead to disentanglement.
}
\end{subfigure}
\begin{subfigure}[h]{0.49\textwidth}
\includegraphics[width=\linewidth]{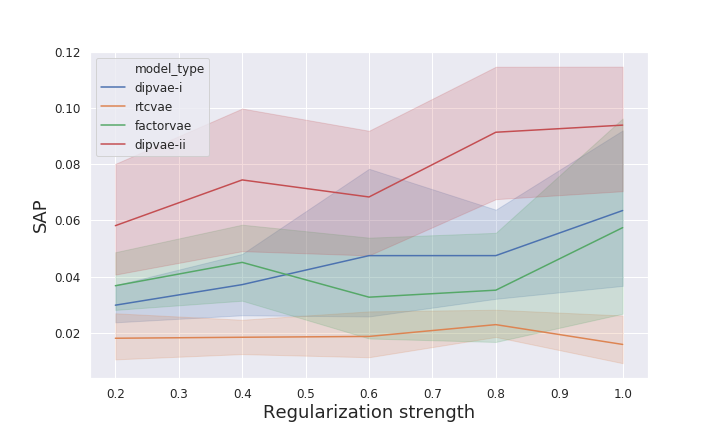}
\caption{The SAP score verses regularization strength on dSprites.}
\end{subfigure}
\begin{subfigure}{.49\textwidth}
\hbox{\hspace{-0.2cm}\includegraphics[width=\linewidth]{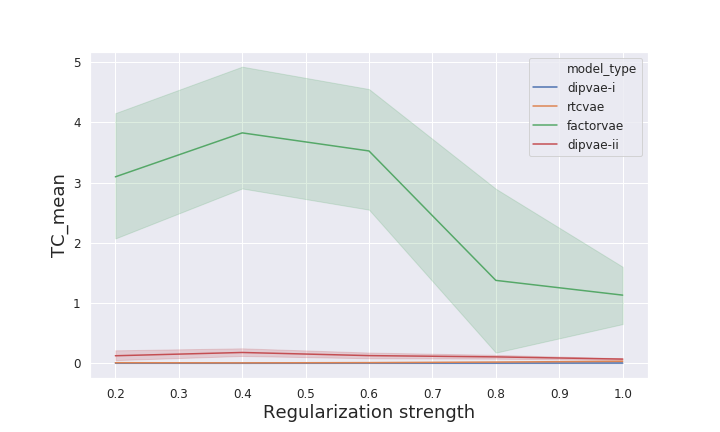}}
\centering
\caption{Direct comparison of $TC_{mean}$ of FactorVAE, DIP-VAEs and RTC-VAE on dSprites. Though both DIP-VAEs and RTC-VAE have low $TC_{mean}$, there is a difference in terms of factorized aggregated posterior.} 
\end{subfigure}

\caption{Scores evaluated on dSprites.}

\label{fig:scores_shape}
\end{figure}

\begin{figure}[h]
\begin{subfigure}{0.49\textwidth}
\includegraphics[width=\linewidth]{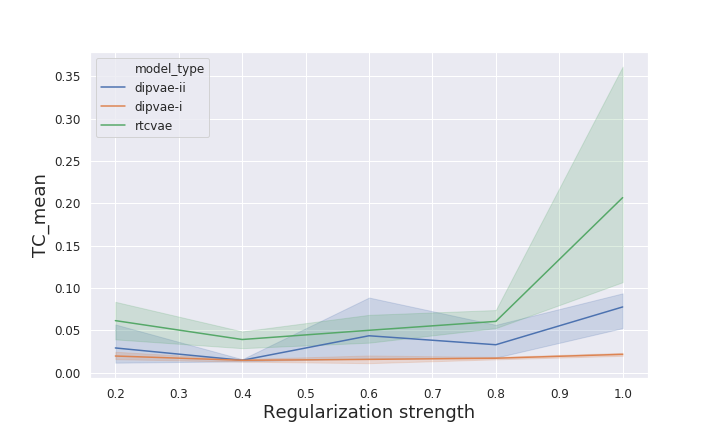}
\caption{$TC_{mean}$ verses regularization strength.}
\end{subfigure}
\hfill
\begin{subfigure}{0.49\textwidth}
\includegraphics[width=\linewidth]{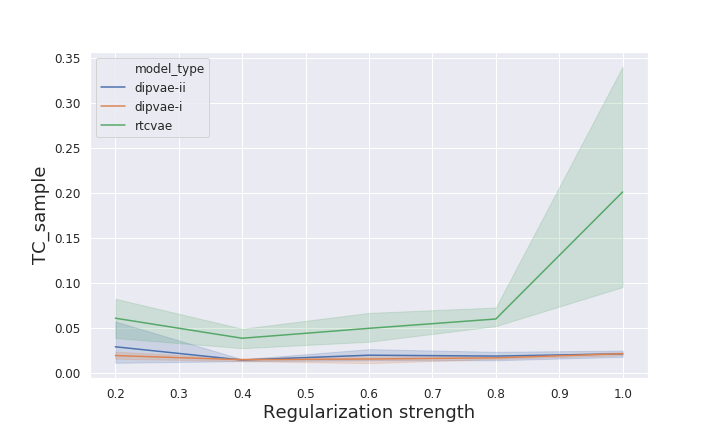}
\caption{$TC_{sample}$ verses regularization strength.}
\end{subfigure}
\begin{subfigure}{0.49\textwidth}
\includegraphics[width=\linewidth]{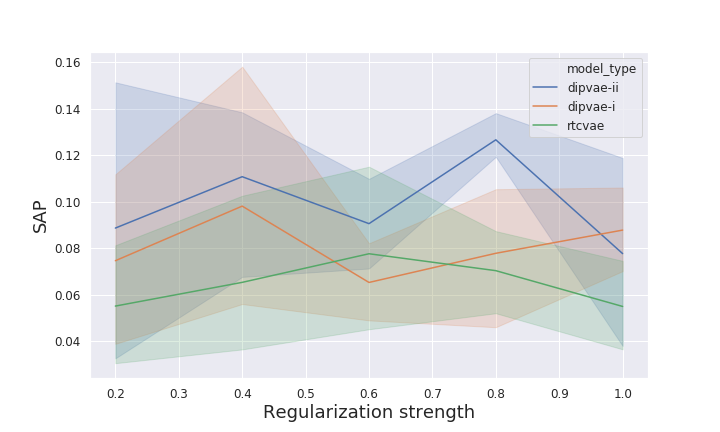}
\caption{The SAP score of models on Car3D under all regularization strength.}
\end{subfigure}
\hfill
\begin{subfigure}{0.49\textwidth}
\includegraphics[width=\linewidth]{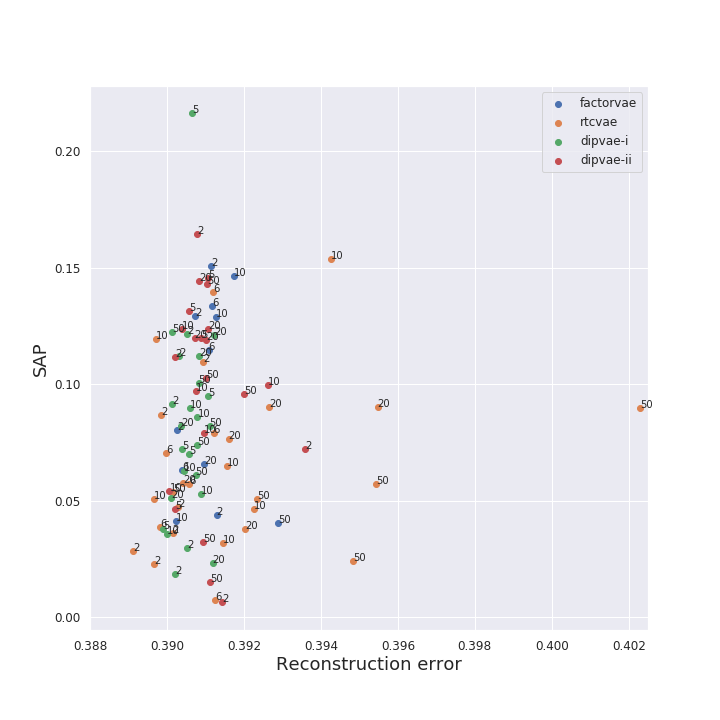}
\caption{SAP score verses reconstruction error.}
\end{subfigure}
\begin{subfigure}{0.49\textwidth}
\includegraphics[width=\linewidth]{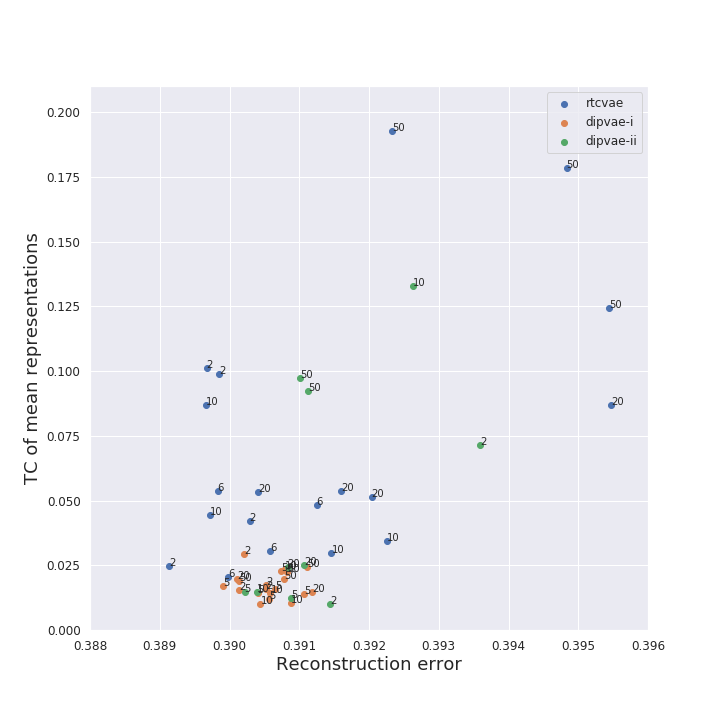}
\caption{$TC_{mean}$ verses reconstruction error. For RTCVAE, $\beta>10$ can affect the quality of reconstruction.}
\end{subfigure}
\hfill
\begin{subfigure}{0.49\textwidth}
\includegraphics[width=\linewidth]{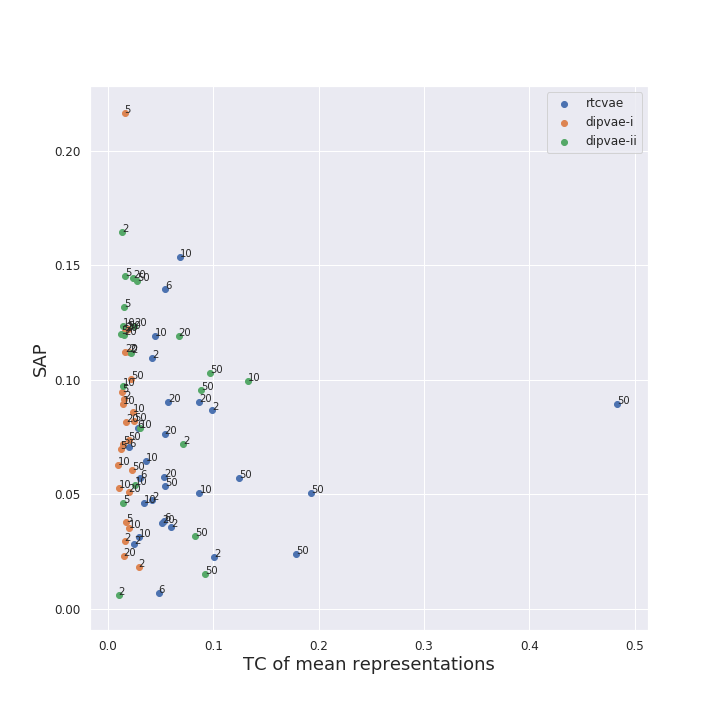}
\caption{Pairplot of SAP score and $TC_{mean}$ shows no strong correlation, indicating factorizing alone does not guarantee disentanglement.}
\end{subfigure}
\caption{Scores evaluated on Car3D.}
    \label{fig:scores_car3d}
\end{figure}

\begin{figure}[h]
\begin{subfigure}[h]{0.45\columnwidth}
\includegraphics[width=\columnwidth]{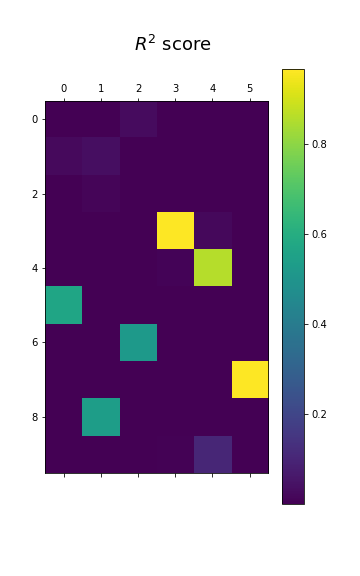}
\end{subfigure}
\hfill
\begin{subfigure}[h]{0.45\columnwidth}
\includegraphics[width=\columnwidth]{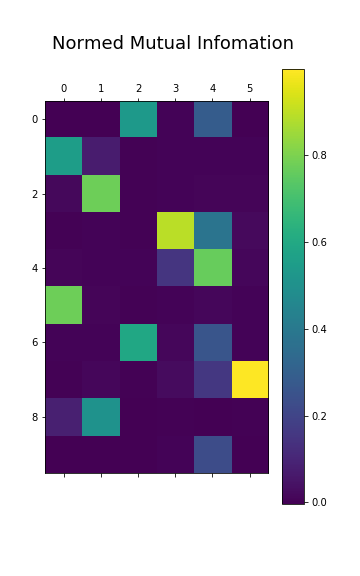}
\end{subfigure}
\hfill
\caption{Evaluating a DIP-VAE-I model with $R^2$ scores and normalized mutual information on Shape3D. 
Though DIP-VAEs do not guarantee factorized representation, they are also affected by non-compactness. (Left) $R^2$ score seems not to show symptom of non-compactness because $R^2$ score only captures the linear relation between random variables; (Right) Mutual information can capture nonlinear relation, so we see a lot more salient values on the right side. E.g., Factor $0,1,2,4$ all have small gaps between top 2 scores.}

\label{fig:R2_MI_s3d_dip}
\end{figure}


\begin{figure}

\begin{subfigure}[t]{0.49\textwidth}
\raisebox{-\height}{\includegraphics[width=\textwidth]{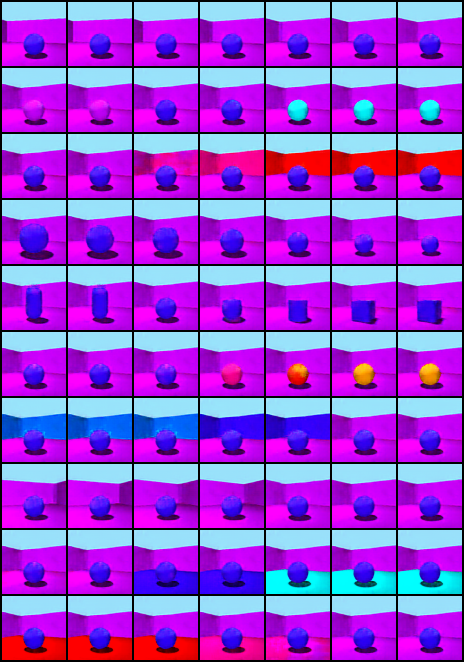}}
\caption{Latent walk of all dimensions of RTC-VAE on Shape3D shows factor disintegration. Orientation: 0, 7; wall hue:2, 6; floor hue:8, 9; object hue:1, 5; shape:4; scale:3.}
\end{subfigure}
\begin{subfigure}[t]{0.49\textwidth}
        \raisebox{-\height}{\includegraphics[width=\textwidth]{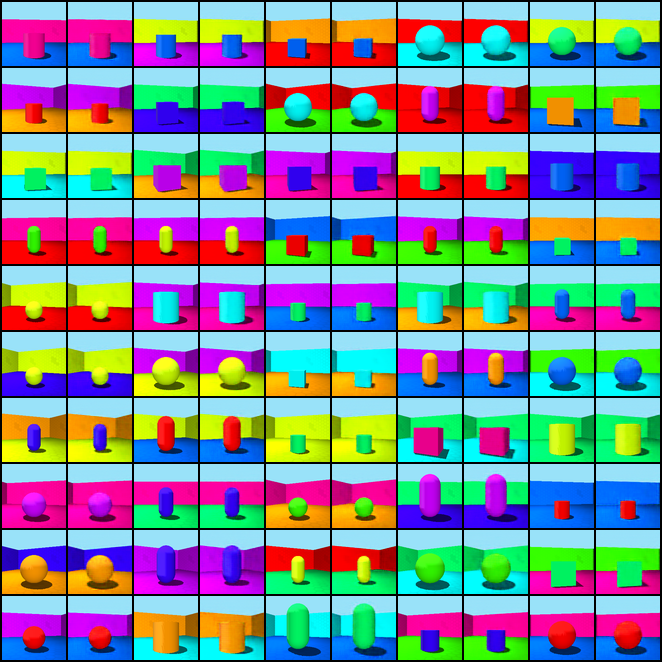}}
        \caption{RTC-VAE reconstruction plot. Every other column is original data, and the next column is reconstruction.}
        \raisebox{-\height}{\includegraphics[width=\textwidth]{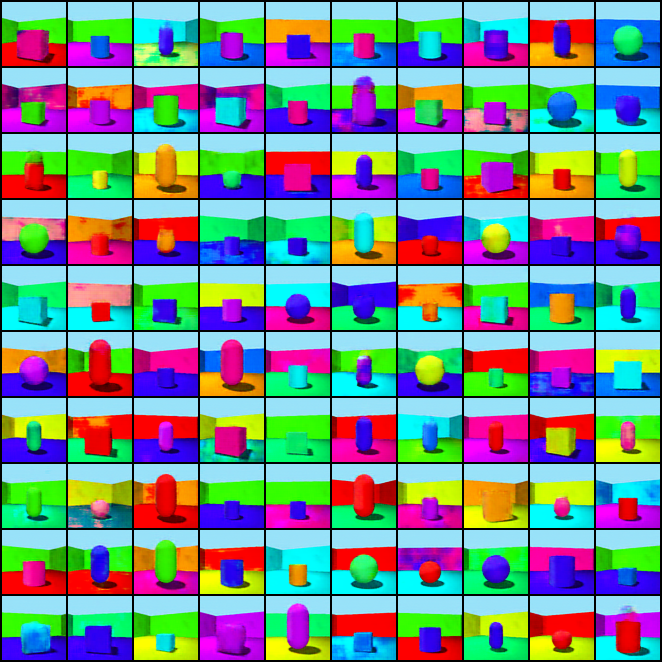}}%
        \caption{Samples from latent space of RTC-VAE.}
    \end{subfigure}
\caption{Experiment results from RTC-VAE with $\beta=10$}
\end{figure}

\begin{figure}[h]
\begin{center}
\includegraphics[scale=0.2]{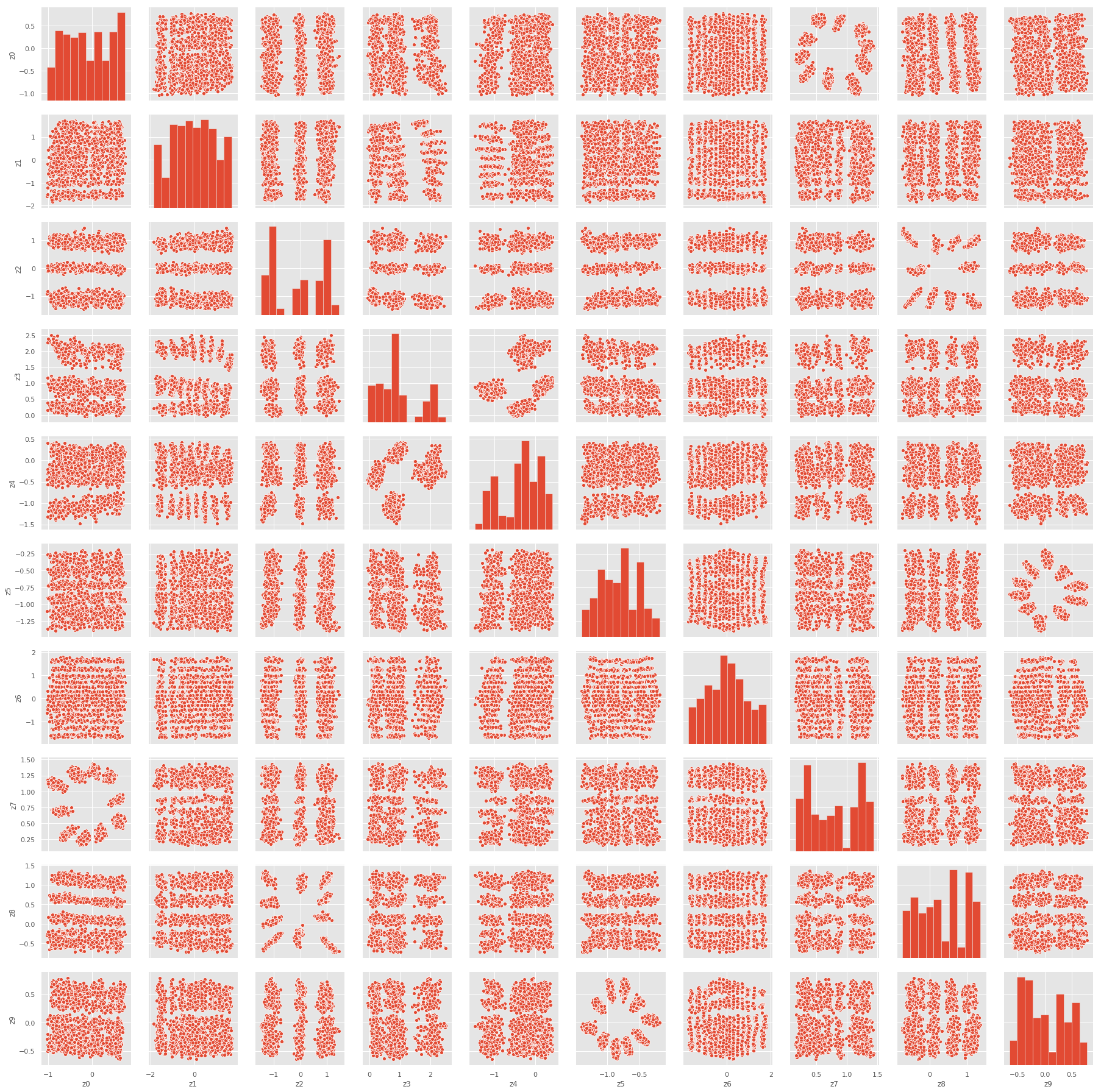}
\centering
\end{center}
\caption{The pairplot of the mean representations of FactorVAE on 2000 samples of Shapes3D, $\gamma=50$. Some dimensions show correlation, e.g. dim 4$\&$3, and some are uncorrelated but not independent, e.g., dim 0$\&$7, dim 2\&8, dim 5\&9 (refer to Section \ref{metrics}).}
\label{fig:factor_s3d_pairplot}
\end{figure}

\begin{figure}[h]
\begin{center}
\includegraphics[scale=0.2]{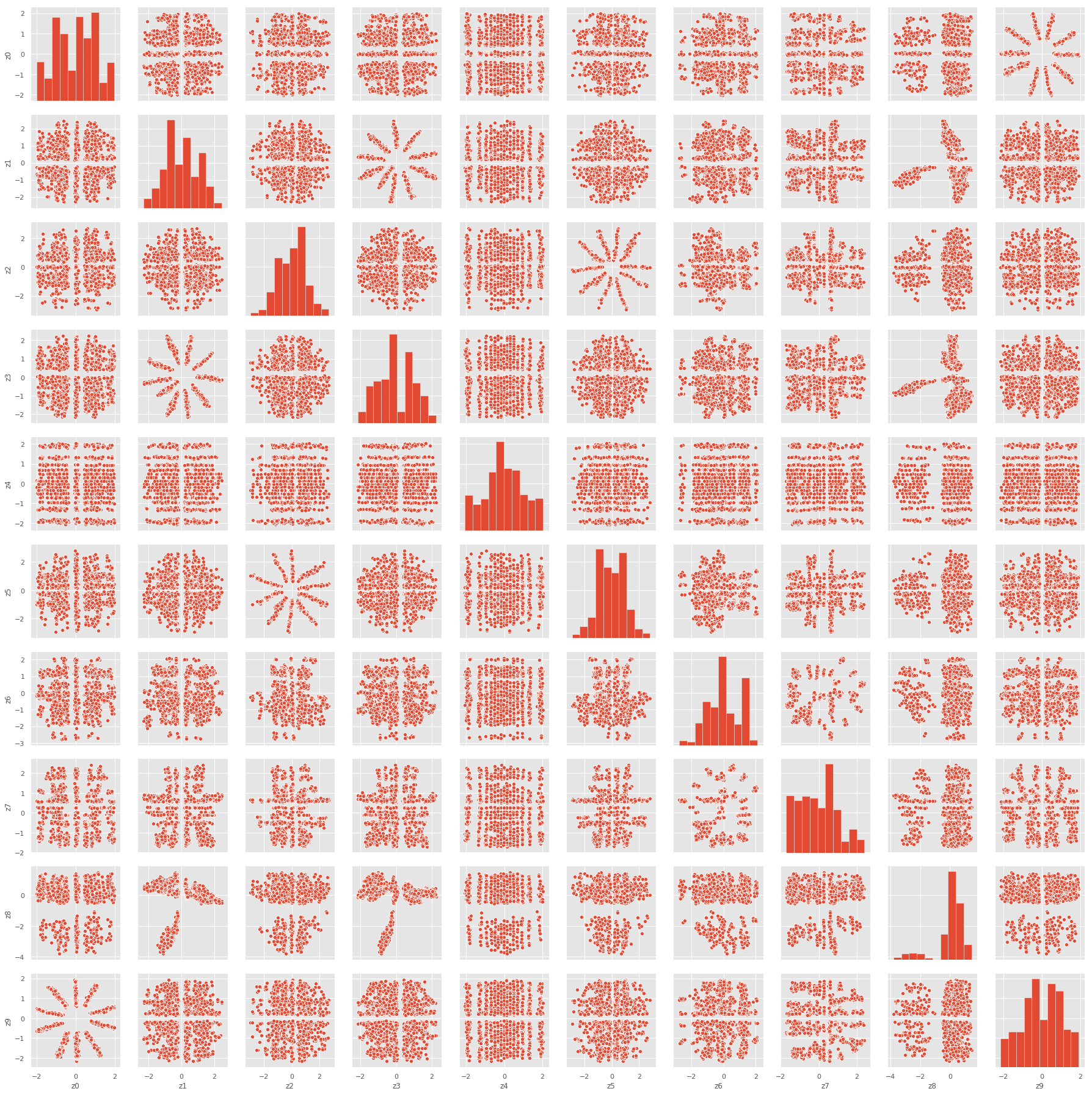}
\centering
\end{center}
\caption{The pairplot of mean representations of DIP-VAE-I on 2000 samples of Shapes3D, $\lambda_{od}=20$. All dimensions are in discrete uncorrelated-like distribution. However, some dimensions are apparently not independent, e.g. dim 1$\&$3, dim 1$\&$3, dim 0$\&$9, dim 2$\&$5, dim 8$\&$1,3, etc (refer to Section \ref{factor_agg_post}).}
\label{fig:DIP_s3d_pairplot}
\end{figure}

\begin{figure}[h]
\begin{center}
\includegraphics[scale=0.2]{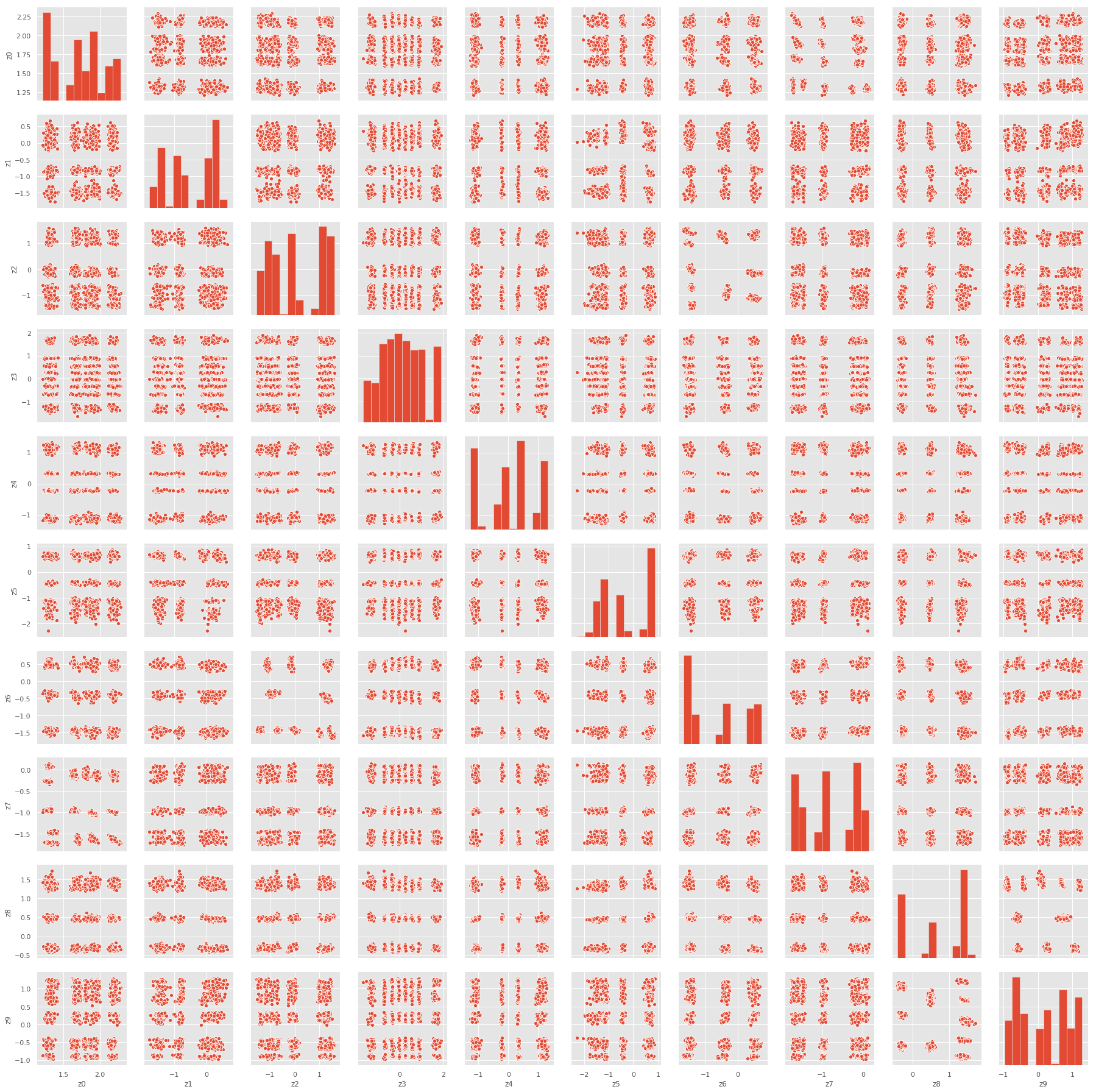}
\centering
\end{center}
\caption{The pairplot of mean reprensentations of RTC-VAE on 2000 samples of Shapes3D, $\beta=20$. All dimensions are in discrete independent-like distribution 
(refer to Section \ref{factor_agg_post}). }
\label{fig:RTC_s3d_pairplot}
\end{figure}
\end{document}